%% file: main.tex
\documentclass[runningheads]{llncs}

\usepackage{eccv}

\usepackage{multirow}
\usepackage{graphicx}
\usepackage{booktabs}
\usepackage{subcaption}
\usepackage[table]{xcolor}
\usepackage{pgfplots}
\pgfplotsset{compat=1.18}
\usepackage{etoc}
\usepackage[accsupp]{axessibility}  %
\usepackage{adjustbox}

\usepackage[pagebackref,breaklinks,colorlinks,citecolor=eccvblue]{hyperref}

\usepackage{orcidlink}

\begin{document}

\title{Enhancing Fine-Grained Spatial Grounding in 3D CT Report Generation via Discriminative Guidance}
\titlerunning{ }

\author{
Chenyu Wang\inst{1} \and
Weicheng Dai\inst{1} \and
Han Liu\inst{2} \and
Wenchao Li\inst{1} \and
Kayhan Batmanghelich\inst{1}
}

\authorrunning{ }

\institute{
Boston University, Boston, MA, USA \\
\email{\{chyuwang, wd2119, wenchao, batman\}@bu.edu}
\and
Siemens Healthineers, Princeton, NJ, USA \\
\email{han.liu@siemens-healthineers.com}
}

\maketitle

\label{sec:abstract}
\input{Sections/main/abstract_v1}

\section{Introduction}
\label{sec:intro}
\input{Sections/main/intro_v2}

\section{Related Work}
\label{sec:related_work}
\input{Sections/main/related_work_v1}

\section{Method}
\label{sec:method}
\input{Sections/main/method_v2}

\section{Experiments}
\label{sec:experiments}
\input{Sections/main/experiments_v2}

\clearpage
\section{Conclusion}
\input{Sections/main/conclusion}
\section{Disclaimer}
For research purposes only. Not for clinical use. This prototype is still under development and not yet commercially available. Future commercial availability cannot be guaranteed.

\newpage
\bibliographystyle{splncs04}
\bibliography{main}

\clearpage
\appendix
\makeatletter
\renewcommand{\theHsection}{appendix.\Alph{section}}
\renewcommand{\theHsubsection}{appendix.\Alph{section}.\arabic{subsection}}
\renewcommand{\theHsubsubsection}{appendix.\Alph{section}.\arabic{subsection}.\arabic{subsubsection}}
\makeatother

\renewcommand{\sectionautorefname}{Section}
\renewcommand{\subsectionautorefname}{Subsection}
\section*{Appendix Contents}
We provide additional details, analyses, and results for our paper in the following sections.
\begin{itemize}
    \item \autoref{sec:appendix:a} presents additional method details.
    \begin{itemize}
        \item \autoref{app:msa} describes the multi-scale attention mechanism.
        \item \autoref{app:dcp2_arch} provides the architecture details of DCP-2.
        \item \autoref{app:prompt} shows the prompt template used for LLM-based parsing.
        \item \autoref{app:question_sets} lists the pre-defined clinical question sets.
    \end{itemize}

    \item \autoref{sec:appendix:b} presents dataset, training, and implementation details.
    \begin{itemize}
        \item \autoref{app:data_impl} summarizes the dataset split and CT pre-processing settings.
        \item \autoref{app:train_impl} summarizes the model architecture, compute environment, and optimization settings.
    \end{itemize}

    \item \autoref{sec:appendix:c} describes the evaluation protocol.
    \begin{itemize}
        \item \autoref{app:eval_18_pathology} details the prior 18-finding evaluation and its consistency with our question-driven protocol.
        \item \autoref{app:class_distribution} reports the class distribution of fine-grained annotations from our evaluation pipeline.
    \end{itemize}

    \item \autoref{sec:appendix:d} reports additional experimental results.
    \begin{itemize}
        \item \autoref{app:additional_dcp_results} provides additional results for DCP-1 and DCP-2 on CT-RATE and RadChest-CT.
    \end{itemize}
\end{itemize}

\section{Method Details}
\label{sec:appendix:a}
\input{Sections/appendix/appendix_a_v2}

\clearpage
\section{Training and Implementation Details}
\label{sec:appendix:b}
\input{Sections/appendix/appendix_b}

\clearpage
\section{Evaluation Protocol}
\label{sec:appendix:c}
\input{Sections/appendix/appendix_c}

\clearpage
\section{Additional Experimental Results}
\label{sec:appendix:d}
\input{Sections/appendix/appendix_d_v2}
\clearpage

\end{document}

%% file: Sections/main/abstract_v1.tex
\begin{abstract}
Vision--language models (VLMs) for radiology report generation (RRG) can produce long-form chest CT reports from volumetric scans and show strong potential to improve radiology workflow efficiency and consistency.
However, existing methods face two key limitations: (i) training supervision is often coarse, aligning a whole CT volume with a full free-text report without explicit alignment for fine-grained attributes or pathology locations; and (ii) evaluation is typically holistic (lexical overlap, entity matching, or LLM-as-a-judge scores) and not diagnostic for spatial grounding.
We propose \emph{Discriminative Cue-Prompting with Prompt Dropout (DCP-PD)}, a plug-and-play framework that distills fine-grained cues from free-text reports and uses them to guide report generation while mitigating shortcut reliance via prompt dropout.
DCP-PD achieves state-of-the-art performance on CT-RATE, improving macro F1 from $=0.501$ to $0.603$ (\(\sim\)20\% relative), and substantially boosts out-of-distribution performance on Rad-ChestCT from F1 $=0.266$ to $0.503$ (\(\sim\)89\% relative).
Finally, we introduce a hierarchical, location-aware question-set protocol (presence $\rightarrow$ laterality $\rightarrow$ lobe) to directly assess pathology-location grounding, showing that fine-grained spatial localization remains challenging even for models that score highly on current benchmarks.
\end{abstract}

%% file: Sections/main/intro_v2.tex
Radiology report generation (RRG) improves clinical workflow efficiency and consistency~\cite{monshi2020deep,tanno2025collaboration}. Clinical accuracy within these generated reports is critical. However, current vision-language models (VLMs) face two primary gaps. First, VLM performance on specific abnormality detection tasks remains inferior to dedicated discriminative models. Second, existing evaluation metrics are too coarse to accurately measure fine-grained spatial grounding. This work introduces a modular approach to address these gaps. We leverage pretrained discriminators to guide VLM generation. Crucially, our framework offers modular extensibility. It allows the discriminator to be updated or replaced without requiring any retraining of the underlying language model.

Previous methods attempt to improve fine-grained alignment through anatomical segmentation~\cite{shui2025large,chen2025large} or auxiliary classifiers~\cite{jin2024promptmrg,wang2023can,chen2025dia}. Segmentation often fails for small, sparse abnormalities. Classifier-based approaches inject diagnostic predictions directly into text prompts to guide generation. Both paradigms share a critical flaw. They tightly couple the VLM to a specific auxiliary model architecture. Consequently, if the auxiliary segmentation or classification model is updated, the entire VLM requires costly retraining.

Standard lexical overlap metrics fail to sufficiently reflect clinical relevance. Recent methods evaluate clinical correctness by parsing reports into specific abnormalities to compute targeted metrics like precision, recall, and F1 scores~\cite{ctchat}. Several 3D CT vision-language works also adopt LLM-as-a-judge evaluations~\cite{lee2024read,ctchat,shui2025large}. While these approaches improve upon lexical scores, they remain too coarse. They are not fine-grained enough to rigorously assess the precise spatial localization of an abnormality. Generating \emph{structured reports} rather than unstructured text can partially address this issue and improve spatial accuracy~\cite{delbrouck2025automated}. Ultimately, a VLM that seamlessly integrates fine-grained discriminative models can directly bridge these localization gaps and substantially improve the clinical accuracy of the generated report.

To achieve modular extensibility, we propose Discriminative Cue-Prompting with Prompt Dropout (DCP-PD). During training, our method extracts ground-truth clinical cues directly from actual free-text reports. We format these cues as text prefixes to condition the VLM. Simply adding ground-truth text can cause the VLM to learn a shortcut, where it merely copies the prompt instead of relying on the image. To prevent this, we develop prompt dropout. This technique randomly drops cue entities during training, forcing the model to verify features against the visual tokens and maintain strong visual grounding. This design fully decouples the VLM from the auxiliary model. At inference time, the prompt is dynamically populated by predictions from a pretrained discriminative model. This strategy is highly modular. We can replace the classifier with a better-performing one at any time without requiring any retraining of the VLM. Finally, this flexibility directly addresses the aforementioned localization shortcomings. It allows us to seamlessly incorporate finer-grained discriminative models that detect not only the presence of an abnormality but also its precise spatial location, thereby improving the overall clinical accuracy of the generated report.

\noindent\textbf{Our contributions are as follows.}
\begin{itemize}
\item We propose DCP-PD, a plug-and-play framework that conditions CT report generation on templated natural-language cues and enables inference-time guidance from arbitrary discriminative models.

\item We train a structured CT report generation VLM with DCP-PD that achieves state-of-the-art performance on CT-RATE across clinical and lexical metrics, and generalizes well to the out-of-distribution Rad-ChestCT benchmark.

\item DCP-PD scales beyond presence/absence of finding labels to include location cues, such as laterality and lobe information for common lung findings.

\item We introduce a diagnostic question-set protocol to evaluate spatial grounding of generated reports hierarchically (presence $\rightarrow$ laterality $\rightarrow$ lobe), and show that fine-grained localization remains challenging.

\end{itemize}

%% file: Sections/main/related_work_v1.tex
\subsection{Fine-Grained Vision-Text Alignment in CT VLMs}
CT vision–language alignment has recently been improved by moving from global scan–report matching to anatomy/region-level alignment. CT-GLIP aligns organ-level 3D CT features with corresponding organ text via grounded contrastive pretraining~\cite{lin2024ct}, while fVLM~\cite{shui2025large} aligns anatomical region features with anatomy-specific report sentences using fine-grained contrastive learning. Reg2RG~\cite{chen2025large} and MedRegion-CT~\cite{kyung2025medregion} incorporate region-centric tokens to condition report generation on anatomical regions. However, these approaches largely rely on anatomy segmentation models~\cite{wasserthal2023totalsegmentator}, so the alignment remains organ/anatomy-level and can be hard to scale to finer-grained or more flexible concepts (e.g., severity, morphology/shape) and small findings where segmentation is noisy or unreliable. 
In contrast, we decouple fine-grained concept alignment from the VLM by training concept-specific discriminators on multi-scale frozen CT embeddings with report-distilled labels, and inject their outputs as cue prompts into the VLM, enabling flexible extension to new concepts.
\subsection{Prompt-Guided Report Generation}
Prompt learning has also been explored for radiology report generation. Wang et al.~\cite{wang2023can} design disease-enriched prompts and further propose an automatic prompt learning mechanism to reduce manual prompt engineering.
Building on this direction, PromptMRG~\cite{jin2024promptmrg} converts predictions from an auxiliary disease classifier into diagnosis-driven textual prompts to steer report generation, and Dia-LLaMA~\cite{chen2025dia} injects diagnostic guidance into LLM-based generation to emphasize clinically salient diseases. However guidance cues from those methods are typically produced by auxiliary models and are largely restricted to predefined presence/absence of finding labels, making them tightly coupled to specific classification architectures. 
In contrast, we use ground-truth supervision at training time to align diverse, fine-grained cues extracted from free-text reports with multi-scale CT embeddings, enabling richer 
cue types during training.
\subsection{Evaluation of Radiology Report Generation}
Radiology report generation is commonly evaluated with lexical overlap metrics~\cite{papineni2002bleu,banerjee2005meteor}, but these correlate weakly with clinical correctness and often miss factual errors~\cite{yu2023evaluating}. 
To better capture clinical content, RadBERT~\cite{ctchat} uses a BERT classifier to compute Micro/Macro F1 over 18 abnormalities from generated and ground-truth reports. 
LLM-based evaluators further assess faithfulness: RadFact~\cite{bannur2024maira} measures bidirectional sentence-level support between generated and reference reports, while GREEN~\cite{ostmeier2024green} provides clinically grounded error annotations for more interpretable evaluation. 
However, these metrics remain holistic and are not diagnostic of pathology--location grounding. 
We therefore complement them with a grounding-oriented protocol that distills location-aware pathology labels from generated and ground-truth reports for fine-grained spatial analysis.

%% file: Sections/main/method_v2.tex
\begin{figure}[!t]
  \centering
  \includegraphics[width=\columnwidth]{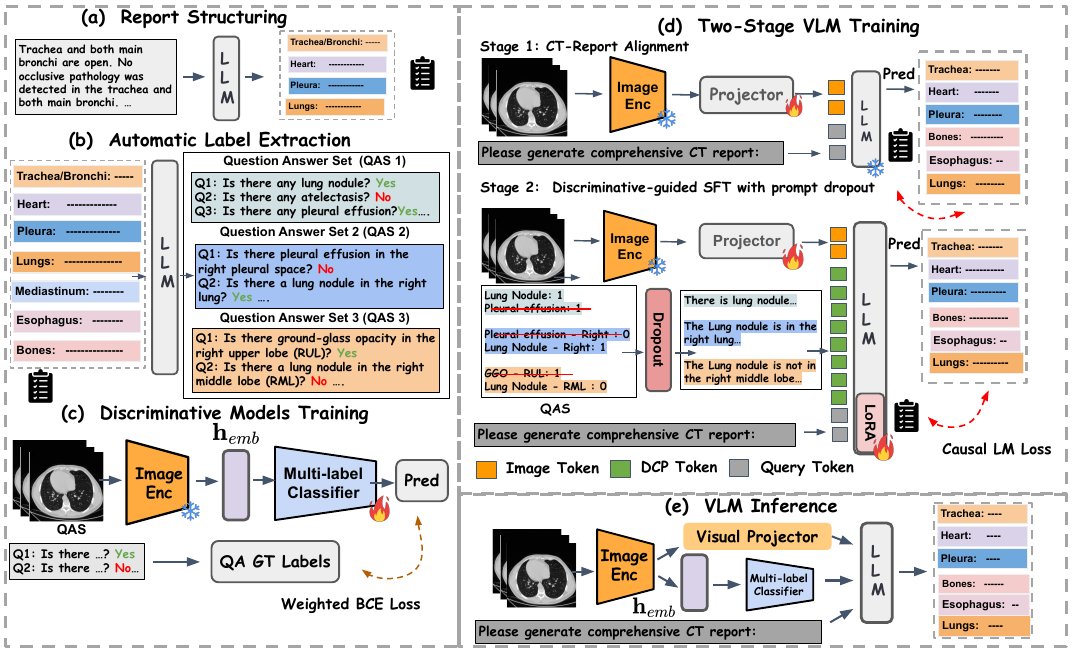}
  \caption{\textbf{Overview of the data preparation, training, and inference pipeline of the DCP-PD framework.}
\textbf{(a) Report structuring:} we convert free-text CT reports into sectioned, anatomy-aware text using an \textsc{LLM}.
\textbf{(b) Automatic label extraction:} we define three set of location-aware clinical questions, and use the same \textsc{LLM} to extract binary answers from each structured report, forming question--answer sets (QAS1--3). Here, QAS1 targets finding-level presence/absence, while QAS2--3 progressively probe finer localization.
\textbf{(c) Discriminative model training:} using Multi-scale CT embeddings from the frozen image backbone, we train lightweight multi-label classifiers with weighted BCE to predict the QA-derived labels, with one classifier trained per question set.
\textbf{(d) Two-stage VLM training:} the image encoder is frozen throughout. Stage~1 performs CT Image--Report alignment by tuning the visual projector for report generation. Stage~2 performs discriminative-guided SFT by prepending ground-truth (GT) cue prompts with prompt dropout and jointly tuning the projector and a LoRA-tuned \textsc{LLM} with a causal LM loss.
\textbf{(e) Inference:} the trained classifier predicts cue prompts, which are injected through the same prompt interface alongside image tokens (and query tokens) to guide report generation.
  }
  \vspace{-0.3em}
  \label{fig:pipeline}
\end{figure}



Our goal is to develop a modular framework for chest CT report generation that supports using discriminative cues to guide a vision-language model toward clinically accurate reports. To achieve this, the framework should satisfy three key requirements: (1) it should provide finer-grained supervision than free-text reports, (2) it should remain decoupled from any specific auxiliary discriminator and enable inference-time guidance from arbitrary discriminative models, and (3) it should naturally extend to finer-grained clinical cues, including both object presence and spatial attributes. Motivated by these requirements, we introduce two key designs.
(a) Report structuring and Question-driven extraction for fine-grained supervision:
we reformulate free-form CT reports into structured, anatomy-aware representations, and use predefined clinical questions to distill binary labels from the structured reports. This yields stronger supervision for both abnormality presence and spatial attributes. (b) Language-based cue interface for modular guidance: we define a unified natural-language cue prompt interface. During training, ground-truth labels extracted from reports are templated into cue prompts to condition the VLM. At inference time, the same interface is populated by predictions from a pretrained discriminative model. Because the VLM is trained on the templated format rather than on any specific model's outputs, it remains decoupled from the discriminator and allows different discriminative models to be swapped in at inference time without retraining the generator. However, training with ground-truth cue prompts may induce shortcuts, where the model copies cue text instead of grounding predictions in the visual input. To mitigate this issue, we apply \emph{prompt dropout} during training by randomly removing cue entities, encouraging the model to rely on visual inputs when cues are incomplete. Together, (a) and (b) naturally satisfy (3): the same framework can extend to finer-grained clinical cues by adding new questions and templating their outputs, without modifying the generator architecture. Fig.~\ref{fig:pipeline} illustrates the data preparation, training, and inference pipeline, and Fig.~\ref{fig:inference} shows the model architecture and test-time flow.
\subsection{Data Pre-processing}
\noindent\textbf{Structured Reports.}
We utilize RadExtract~\cite{goel_radextract_2025} to convert raw, unstructured chest CT reports into structured, anatomy-aware representations as shown in Fig.~\ref{fig:pipeline}(a). 
RadExtract applies prompt-driven extraction to separate report metadata, clinical findings, and impression content, and organizes findings into section-labeled units with clinical-importance annotations. 
We run the pipeline with the LLM where few-shot exemplars and explicit JSON constraints in the prompt encourage consistent outputs and robust handling of common edge cases.

\noindent\textbf{CT Pre-processing.}
To handle variability in CT resolution and scan length across studies, we (1) resample the volume to a standardized voxel spacing, (2) normalize axial extent by taking a centered crop for longer scans and symmetric center-padding for shorter scans, (3) apply center crop/pad to obtain a fixed in-plane resolution, and (4) To enhance the contrast of different dominant tissue types in chest CT (e.g., lung, mediastinum and bone), we further apply multiple contrast windows, as shown in Fig.~\ref{fig:inference}. Each window corresponds to a specific mapping from Hounsfield Units~\cite{denotter2019hounsfield} to intensity values, tailored to the attenuation range of a particular tissue type, yielding an 11-channel multi-window CT input.


\subsection{Visual Backbone and Vision Projector}
\begin{figure}[t!]
  \centering
  \includegraphics[width=\columnwidth]{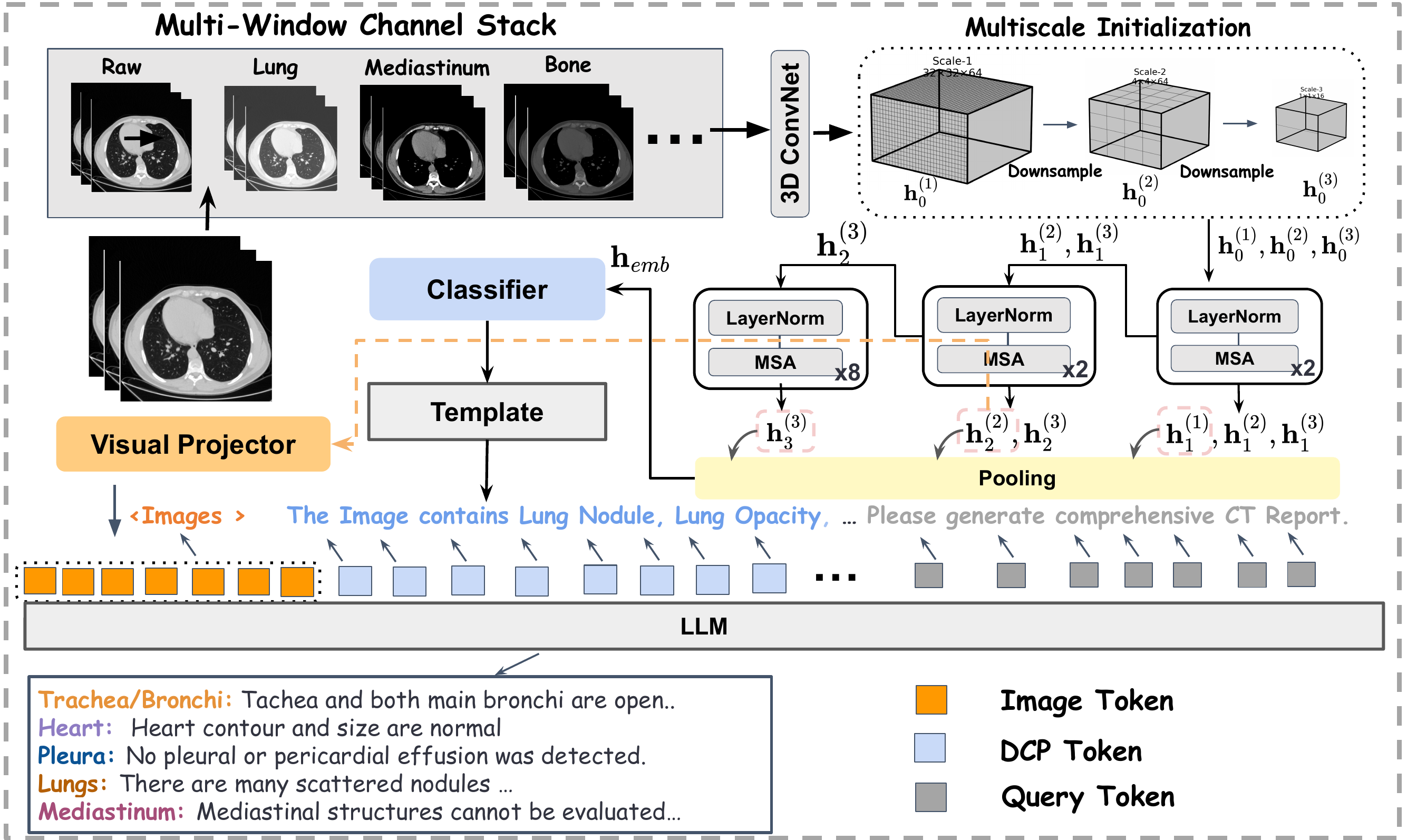}
  \caption{\textbf{Detailed inference schematic.}
The pre-processed multi-channel CT volume is encoded by a 3D convolutional stem to produce a fine-scale token grid, followed by progressive downsampling to construct multi-scale token hierarchies. These token grids are processed by the Atlas visual backbone with stacked MSA blocks to yield hierarchical representations. The aggregated embedding $h_{\mathrm{cls}}$ is fed into a linear classifier to produce DCP tokens via templating, while selected intermediate Atlas tokens $h^{(2)}_{2}$ are projected into LLM image tokens through a vision projector. The LLM generates the final report conditioned on the concatenation of image tokens, DCP tokens, and query tokens.}
  \vspace{-0.8em}
  \label{fig:inference}
\end{figure}

\noindent\textbf{Visual Backbone.}
We use the Pillar-0 foundation model with its Atlas vision backbone, built on Multi-Scale Attention (MSA) for efficient long-context image modeling~\cite{pillar0,agrawal2025atlas}. Fig.~\ref{fig:inference} illustrates the multi-scale token hierarchy and how Atlas processes CT volumes with stacked \textsc{MSA} blocks.
Given a pre-processed chest CT volume $\mathbf{V}\in\mathbb{R}^{C\times H\times W\times D}$ (with $C$ input channels, in-plane size $H\times W$, and axial depth $D$), Atlas applies a 3D convolutional patchification stem with patch size $\mathbf{p}=(p_H,p_W,p_D)$ to obtain a fine-scale token grid $\mathbf{h}^{(1)}\in\mathbb{R}^{C_{\mathrm{in}}\times \frac{H}{p_H}\times \frac{W}{p_W}\times \frac{D}{p_D}}$, where $C_{\mathrm{in}}$ denotes token embedding dimension.
Atlas constructs a multi-scale hierarchy $\{\mathbf{h}^{(\ell)}\}_{\ell=1}^{L}$ via iterative strided pooling: for $\ell=2,\dots,L$, $\mathbf{h}^{(\ell)} \leftarrow \mathrm{StridedPooling}(\mathbf{h}^{(\ell-1)}, s_H, s_W, s_D)$, yielding
\begin{equation}
\mathbf{h}^{(\ell)} \in \mathbb{R}^{C_{\mathrm{in}}\times
\frac{H}{p_H s_H^{\ell-1}}\times
\frac{W}{p_W s_W^{\ell-1}}\times
\frac{D}{p_D s_D^{\ell-1}}}, \quad \ell=1,\dots,L.
\end{equation}
\noindent Atlas groups \textsc{MSA} blocks into multiple encoder stages and applies progressive scale dropping for efficiency~\cite{agrawal2025atlas}.
To track stage outputs, we denote the token grid at scale $\ell$ after stage $u$ by $\mathbf{h}^{(\ell)}_{u}$, with $\mathbf{h}^{(\ell)}_{0}$ indicating the hierarchy before any stage updates.
Instead of using the final coarsest output, we extract an intermediate representation for training our vision-language models and use $\mathbf{h}^{(2)}_{2}$, which provides a favorable trade-off between computational cost and visual detail for chest CT report generation.
Full details of \textsc{MSA} are provided in Appendix~A.

\noindent\textbf{Vision Projector.}
To interface Atlas visual tokens with the \textsc{LLM}, we flatten the selected Atlas token grid into a token sequence and apply cross-attention with a small set of learnable query tokens to pool information from the visual tokens into a fixed-length embedding set.
We then map these embeddings to the \textsc{LLM} hidden size using a lightweight two-layer \textsc{MLP} with a \textsc{GELU} nonlinearity~\cite{hendrycks2016gaussian}, and use the resulting outputs to condition the \textsc{LLM}.
\subsection{Discriminative Modules}
\noindent\textbf{Automatic label extraction.}
To obtain supervision beyond coarse report-level alignment, we adopt a \emph{question-driven} extraction interface.
Given a structured report, we prompt an \textsc{LLM} with a predefined set of binary clinical questions and distill yes/no answers, yielding report-derived labels for findings and attributes as illustrated in Fig.~\ref{fig:pipeline}(b).
This interface decouples label specification from the model: extending supervision to new concepts only requires adding or editing questions, without changing the architecture.
We focus on a three-level diagnostic hierarchy that progresses from finding detection to increasingly fine-grained spatial grounding.
Specifically, QAS1 probes finding-level presence/absence, QAS2 queries laterality (left/right), and QAS3 further refines localization to lung lobes.

\noindent\textbf{Multi-scale CT embedding.}
To train the discriminative classifier, we construct a multi-scale CT embedding by pooling and concatenating frozen Atlas outputs across stages:
\begin{equation}
\mathbf{h}_{emb}=\mathrm{Concat}\!\Big[\mathrm{Pool}(\mathbf{h}^{(1)}_{1}),\ \mathrm{Pool}(\mathbf{h}^{(2)}_{2}),\ \ldots,\ \mathrm{Pool}(\mathbf{h}^{(L)}_{L})\Big]\in\mathbb{R}^{LC_{\mathrm{in}}}.
\end{equation}
Here, $\mathrm{Pool}(\cdot)$ aggregates over tokens to produce a $C_{\mathrm{in}}$-dimensional feature.

\noindent\textbf{Discriminative models.}
To obtain discriminative cues, we apply a lightweight multi-label linear probe to the multi-scale embedding $\mathbf{h}_{emb}$ to predict the binary labels for each predefined question set.

\subsection{Discriminative Cue Guided Report Generation}
To generate radiology reports from chest CT volumes, we condition the \textsc{LLM} on a single input sequence formed by concatenating (i) projected image tokens ($\mathbf{T}_{\mathrm{img}}$), (ii) a templated prompt that summarizes predicted findings ($\mathbf{T}_{\mathrm{disc}}$), and (iii) a textual instruction ($\mathbf{T}_{\mathrm{qry}}$). Concretely, 
Given $\mathbf{P}=\big[\,\mathbf{T}_{\mathrm{img}} \,;\, \mathbf{T}_{\mathrm{disc}} \,;\, \mathbf{T}_{\mathrm{qry}} \,\big]$, the \textsc{LLM} decodes the report autoregressively.

\subsection{Training and Evaluation}

\noindent\textbf{Discriminative Models Training.}
We train a lightweight multi-label linear probe on the multi-scale embedding $\mathbf{h}_{emb}$ using a weighted binary cross-entropy objective to mitigate class imbalance.

\noindent\textbf{VLM Training.}
We train our model in two stages and keep the visual backbone frozen throughout.
\textbf{Stage 1 (CT-report alignment).}
We train only the vision projector with the \textsc{LLM} frozen, aligning projected visual tokens to the \textsc{LLM} token space.
\textbf{Stage 2 (Discriminative-guided SFT with prompt dropout).}
We perform parameter-efficient SFT by jointly training the vision projector and LoRA~\cite{hu2022lora} adapters in the \textsc{LLM}. To prevent the LLM from engaging in shortcut learning by copying discriminative cues from $\mathbf{T}_{\mathrm{disc}}$, we apply \emph{prompt dropout} during training (Fig~\ref{fig:pipeline}(d) stage 2).
Prompt dropout is applied to the classifier-derived labels before templating: each predicted finding/attribute entity is independently dropped with probability $p$, and the remaining entities are converted into a textual prompt for conditioning. 

\noindent\textbf{Evaluation.}
To better understand the capability of our discriminative-guided \textsc{VLM} and its spatial grounding ability. We reuse the same \textsc{LLM}-based question--answer extraction pipeline: we parse each reference report once to cache ground-truth answers, and at evaluation time we parse generated reports under the identical question set to compute agreement.
The full question set is provided in Appendix~\ref{sec:appendix:a}.

%% file: Sections/main/experiments_v2.tex
In this section, we conduct experiments to answer the following research questions (RQs). 

\noindent\textbf{RQ1:} How do discriminative models compare to a base VLM across different finding detection tasks?

\noindent\textbf{RQ2:} Does DCP-PD boost the Base VLM on report generation quality and clinical correctness?

\noindent\textbf{RQ3:} Is DCP-PD effective in mitigating shortcuts?

\noindent\textbf{RQ4:} How does our DCP-PD trained VLM perform against state-of-the-art baselines on report generation benchmarks?

\noindent\textbf{RQ5:} Can DCP-PD incorporate richer cue types and improve fine-grained localization?

\noindent Additional implementation details and results are in Appendix~\ref{sec:appendix:b} and Appendix~\ref{sec:appendix:d}. We also present qualitative case studies in Fig.~\ref{fig:case_study} and Fig.~\ref{fig:full_report_vis}.

\subsection{Experimental Setup}
\label{sec:setup}
\noindent\textbf{Datasets.} We use CT-RATE~\cite{hamamci2026generalist} as the primary dataset for report generation, and Rad-ChestCT~\cite{draelos2020rad} for external validation.
For structured report generation and efficient question-set label distillation, we convert the entire CT-RATE corpus from raw free-text into structured, sectioned reports using RadExtract and train our VLM on this structured version.
Additional dataset and preprocessing details are provided in the Appendix~\ref{sec:appendix:b}.

\noindent\textbf{Models.}
We run RadExtract with Gemini-2.5-Flash~\cite{comanici2025gemini} to convert raw CT-RATE reports into structured, sectioned reports, Qwen-30B~\cite{qwen3technicalreport} to parse generated and ground-truth reports into structured answers, and LLaMA-3 Instruct 8B~\cite{liu2023visual} for report generation.
Additional implementation, model, and training details are provided in Appendix~B.

\noindent\textbf{Evaluation Metrics.}
We evaluate report generation quality using standard lexical metrics,  including the mean BLEU score over (BLEU-1--4) and METEOR\cite{papineni2002bleu,banerjee2005meteor}.
For clinical correctness, following prior work, we report Precision, Recall, and F1 on 18 pathology findings, as well as the distribution-aware CRG~\cite{hamamci2025crg} score. In addition, to assess fine-grained grounding ability, we evaluate Precision, Recall, and F1 on location-aware question sets, including laterality- and lobe-specific labels for common lung pathologies.
Details of the evaluation protocol, question-sets are provided in Appendix~C.

\noindent\textbf{Attention Reliance Score.}
To quantify how much the decoder relies on discriminative cue tokens versus image tokens during generation, we compute an attention-based reliance score from self-attention weights.
For each decoding step $t\in\{1,\dots,T\}$, let $\mathbf{a}_t\in\mathbb{R}^{S}$ denote the attention distribution from the newly generated token at step $t$ to all $S$ prompt positions, averaged over all heads and layers.
Let $R_{\mathrm{text}}$ and $R_{\mathrm{image}}$ denote the subsets of prompt positions corresponding to DCP tokens and image tokens, respectively.
We define the region-averaged attention mass for a token region $R_r$ as
\begin{equation}
\bar m_r=\frac{1}{T}\sum_{t=1}^{T}\left(\frac{1}{|R_r|}\sum_{i\in R_r}\mathbf{a}_t(i)\right).
\end{equation}
We then compute the normalized reliance scores as
\begin{equation}
S_{\mathrm{text}}=\frac{\bar m_{\mathrm{text}}}{\bar m_{\mathrm{text}}+\bar m_{\mathrm{image}}},
\qquad
S_{\mathrm{image}}=1-S_{\mathrm{text}}.
\end{equation}
A larger $S_{\mathrm{text}}$ indicates stronger reliance on DCP tokens, while a larger $S_{\mathrm{image}}$ indicates stronger reliance on image tokens.

\subsection{Main Results}
\label{sec:disc_models}

\begin{figure}[!t]
  \centering
  \includegraphics[width=\columnwidth]{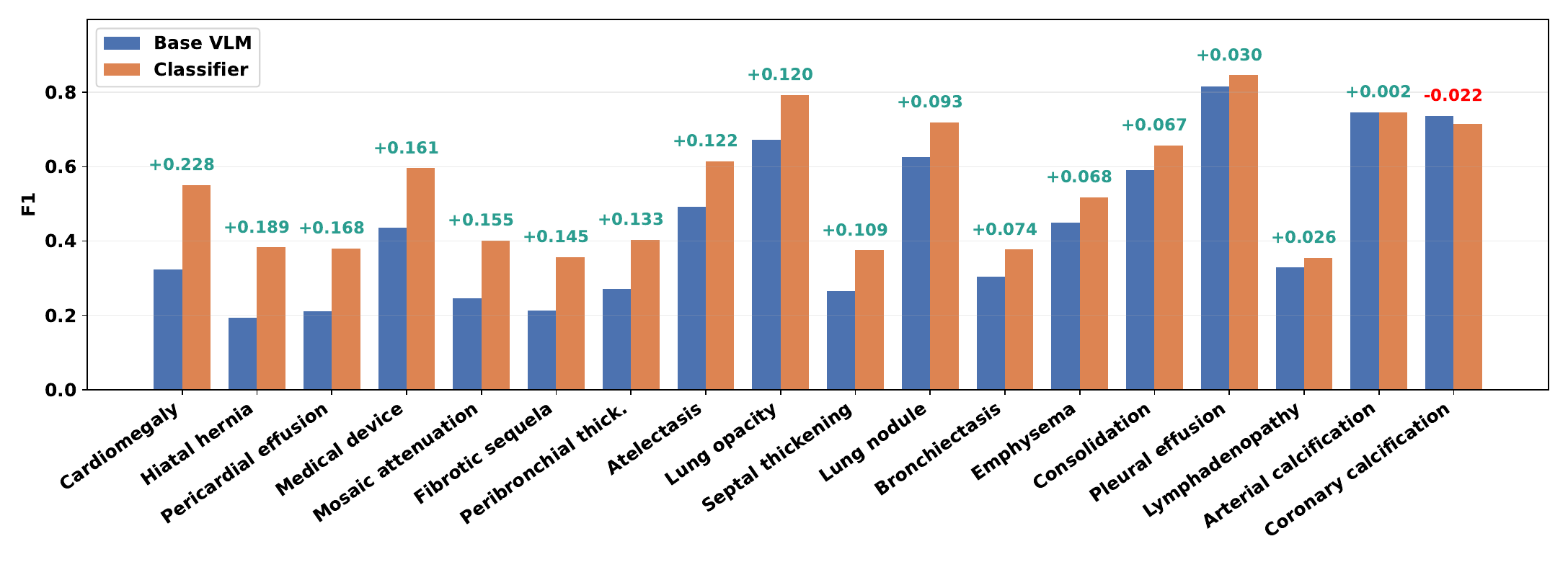}
  \caption{\textbf{Per-pathology detection performance.}
We report F1 for the Base VLM and a linear discriminative classifier on 18 CT findings.
Numbers above bars indicate $\Delta$F1 (Classifier$-$Base VLM). For readability, we shorten several long finding names on the x-axis: Arterial calcification (arterial wall calcification), Coronary calcification (coronary artery wall calcification), Septal thickening (interlobular septal thickening), Mosaic attenuation (mosaic attenuation pattern), Peribronchial thick.\ (peribronchial thickening), and Fibrotic sequela (pulmonary fibrotic sequela).}
  \label{fig:f1_barplot_disc_vs_vlm}
\end{figure}

\begin{table}[t!]
\centering
\caption{Report generation performance under different test-time DCP variants. Models are trained without DCP or with our GT DCP-PD and evaluated with different test-time cue sources. DCP-PD consistently improves report quality and is robust to the choice of test-time DCP source, with additional gains when using stronger cue predictors.}
\label{tab:rq2_dpdcp}
\setlength{\tabcolsep}{1.6pt}
\renewcommand{\arraystretch}{1.05}
\begin{tabular}{llcccccc}
\toprule
Training setting & Test setting & Prec. $\uparrow$ & Recall $\uparrow$ & F1 $\uparrow$ & CRG $\uparrow$ & BLEU $\uparrow$ & METEOR $\uparrow$ \\
\midrule
\textbf{w/o DCP} & No DCP & 0.520 & 0.423 & 0.440 & 0.457 & 0.293 & 0.214 \\
\midrule
\multirow{2}{*}{\textbf{w/ GT DCP-PD}} 
&  DCP-1    & 0.466 & 0.681 & 0.530 & 0.504 & 0.305 & 0.222 \\
& DCP-2 &  \textbf{ 0.540} &\textbf{ 0.700 }&\textbf{ 0.603 }& \textbf{0.555} & \textbf{0.306} & \textbf{0.226}  \\
\bottomrule
\end{tabular}
\end{table}

\begin{table}[h!]
\centering
\caption{Effect of cue prompts dropout training on shortcuts. We compare VLMs trained with GT DCP and GT DCP-PD and evaluate them under the \textit{No DCP} test setting, where cue prompts are removed at inference.
The shaded row highlights the shortcut failure case of GT DCP, which collapses when cue prompts are absent.}

\label{tab:rq3_dropout}
\setlength{\tabcolsep}{1.6pt}
\renewcommand{\arraystretch}{1.05}
\begin{tabular}{llcccccc}
\toprule
Training setting & Test setting & Prec. $\uparrow$ & Recall $\uparrow$ & F1 $\uparrow$ & CRG $\uparrow$ & BLEU $\uparrow$ & METEOR $\uparrow$ \\
\midrule
\rowcolor{red!10}\textbf{w/ GT DCP} & No DCP & 0.393 & 0.071 & 0.094 & 0.349 & 0.268 & 0.200 \\
\midrule
\textbf{w/ GT DCP-PD} & No DCP  & 0.433 & \textbf{0.373} & \textbf{0.366} & 0.436 & 0.288 & 0.211 \\
\bottomrule
\end{tabular}
\end{table}

\begin{table}[!]
\caption{Text and image attention reliance score under the \textit{DCP-2} test-time setting. We report mean\(\pm\)std over 500 validation samples on CT-RATE.
}
\label{tab:attn_reliance}
\centering
\small
\setlength{\tabcolsep}{6pt}
\begin{tabular}{lcc}
\hline
Training setting & $\boldsymbol{S_{\mathrm{text}}}$ (mean$\pm$std) & $\boldsymbol{S_{\mathrm{image}}}$ (mean$\pm$std) \\
\hline
\textbf{w/ GT DCP} & $0.441 \pm 0.054$ & $0.559 \pm 0.054$ \\
\textbf{w/ GT DCP-PD }& $0.316 \pm 0.036$ & $0.684 \pm 0.036$ \\
\hline
\end{tabular}
\end{table}

\noindent\textbf{(RQ1) Discriminative models outperform the base VLM on CT finding classification.}
We evaluate macro F1 on presence/absence prediction of CT findings under the same image encoder and training data.
As shown in Fig.~\ref{fig:f1_barplot_disc_vs_vlm}, linear-probe discriminative models outperform the base VLM on 17 out of 18 findings, with the remaining category showing comparable performance.
Overall, this leads to a 27\% relative improvement in macro F1 (0.56 vs.\ 0.44) over the base VLM.

\noindent\textbf{(RQ2) DCP-PD improves report generation in a model-agnostic manner.} 
 As shown in Table~\ref{tab:rq2_dpdcp}, training the VLM with our \textit{GT DCP-PD}\footnote{We use a dropout rate of 0.3 in all main-paper experiments; additional dropout ablations are reported in Section~\ref{sec:ablation}.} yields substantial gains over the \textit{No-DCP} baseline and remains effective when replacing oracle cues with predicted cues at test time.
In particular, using cues predicted by our discriminative model (\textit{DCP-1}) improves F1 from 0.440 to 0.530 (\(\sim\)20\% relative), together with consistent improvements in lexical metrics, indicating that better discriminative cues lead to better generation.
Importantly, this prompting-based cue interface is \emph{model-agnostic}: leveraging a stronger discriminative model (\textit{DCP-2})\footnote{See Appendix~B for details.} further boosts performance to F1 \(=0.603\), corresponding to a \(\sim\)37\% relative gain over the \textit{No-DCP} baseline.


\begin{table}[!]
\begin{center}
\caption{Comparison of report generation performance under different settings on the \textbf{CT-RATE} benchmark.
We report clinical accuracy metrics (Prec., Recall, F1, CRG) and natural language generation metrics (BLEU(mean), METEOR).
Best results per column are shown in bold.
(\textit{~$^\dagger$ show results from ~\cite{hamamci2025btb3d}; $^\ddagger$ show results from ~\cite{tian2025feature}}).
}
\label{table:ct_report_results}
\setlength{\tabcolsep}{4.0pt}
\renewcommand{\arraystretch}{1.05}
\resizebox{\columnwidth}{!}{%
\begin{tabular}{lcccccc}
\toprule
Method & Prec.$\uparrow$  & Recall$\uparrow$  & F1$\uparrow$  & CRG$\uparrow$  & BLEU(mean)$\uparrow$  & METEOR$\uparrow$  \\
CT2Rep$^{\dagger}$~\cite{CT2Rep}   & 0.435 & 0.128 & 0.160 & 0.359 & 0.280 & 0.197 \\
Merlin$^{\dagger}$~\cite{blankemeier2024merlin}   & 0.295 & 0.112 & 0.160 & 0.352 & 0.260 & 0.148 \\
CT-CHAT$^{\dagger}$~\cite{ctchat}  & 0.450 & 0.158 & 0.184 & 0.368 & 0.272 & 0.215 \\
BTB3D-16$^{\dagger}$~\cite{hamamci2025btb3d} & 0.260 & 0.260 & 0.258 & 0.370 & 0.305 & 0.223 \\
LLM+S-LMR+CSE$^{\ddagger}$~\cite{tian2025feature}             & 0.468 & 0.166 & 0.189 & - & 0.286 & 0.219 \\ 
MedVInT-TD$^{\ddagger}$~\cite{zhang2023pmcvqa}  & 0.408 & 0.097 & 0.121 & - & 0.252 & 0.184 \\
Med-Flamingo$^{\ddagger}$~\cite{pmlr-v225-moor23a}  & 0.422 & 0.105 & 0.138 & - & 0.260 & 0.191 \\
RadFM$^{\ddagger}$~\cite{wu2023generalist}  & 0.430 & 0.122 & 0.154 & - & 0.267 & 0.199 \\
CT-GRAPH~\cite{kalisch2025ct}    & 0.396 & 0.248 & 0.296 & - & \textbf{0.353} & - \\
MS-VLM~\cite{lee2024read}  & 0.222 & 0.329 & 0.261 & - & -  & - \\
CT-AGRG w/ CT-Net~\cite{di2025ct}  & 0.457 & 0.630 & 0.501 & - & - & 0.196 \\

\midrule
Base VLM~(Ours) & 0.520 & 0.423 & 0.440 & 0.457 & 0.293 & 0.214 \\
DCP-PD - DCP1~(Ours)
& 0.467 & 0.681 & 0.530& 0.504 & 0.305 & 0.222 \\
DCP-PD - DCP2~(Ours)
& \textbf{0.540} & \textbf{0.700} & \textbf{0.603} & \textbf{0.556} & 0.306 & \textbf{0.226} \\
\bottomrule
\end{tabular}%
}
\end{center}
\end{table}

\begin{figure}[!]
  \centering
  \includegraphics[width=0.85\columnwidth]{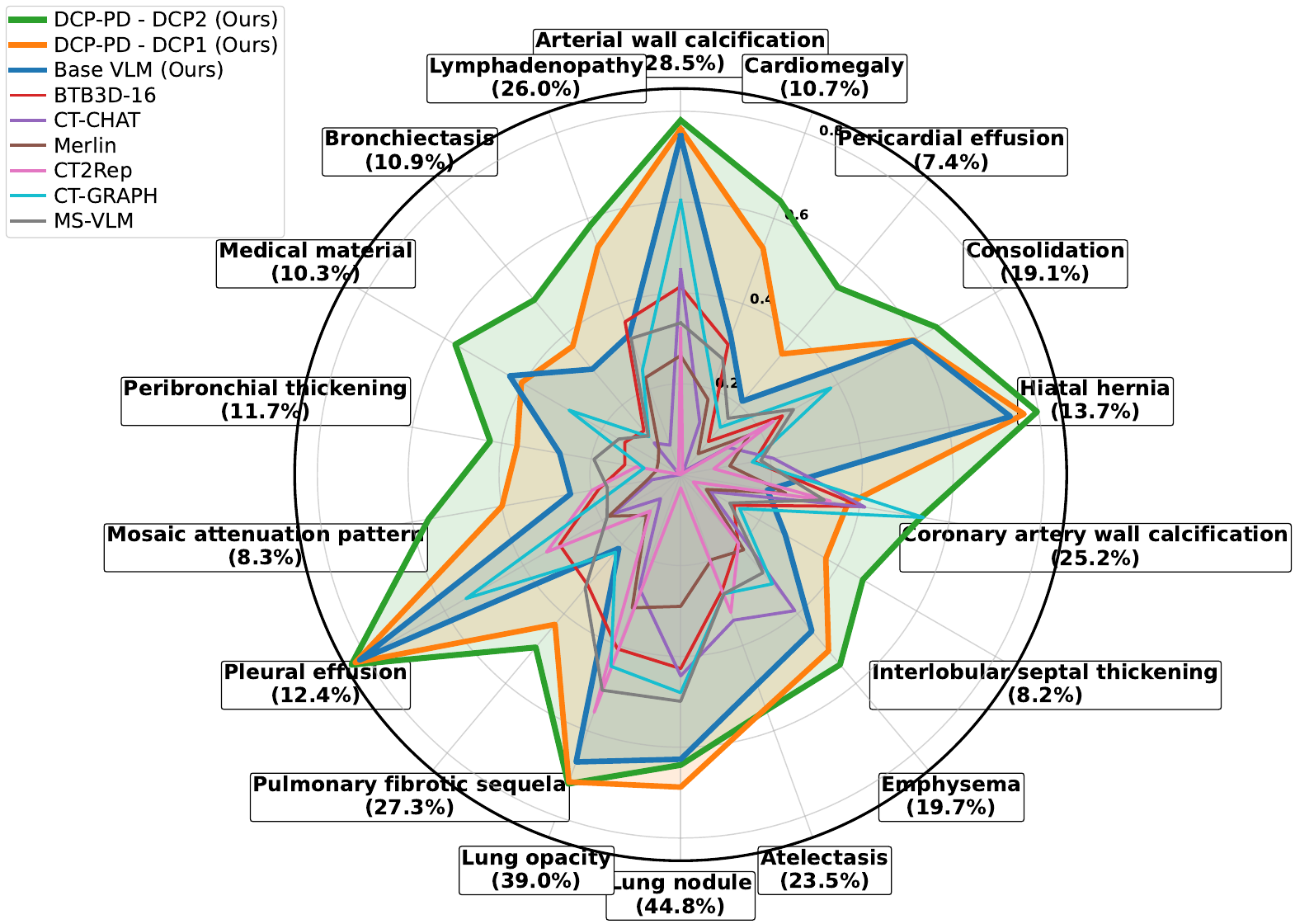}
    \caption{\textbf{Per-pathology performance on CT-RATE.} We compare our \textsc{Base VLM} and a \textsc{DCP-PD}-trained generator under two inference-time guidance settings (\textit{DCP-1} and \textit{DCP-2}) against prior baselines. We report per-pathology F1 for each finding. The value in parentheses next to each label indicates the positive rate (prevalence) in the CT-RATE validation set, and larger radii correspond to higher F1.}
  \label{fig:unstructured_radar}
\end{figure}

\begin{table}[!]
\centering
\caption{Evaluation on \textbf{Rad-ChestCT} demonstrates strong out-of-distribution generalization.
As only binary labels are available, text-based metrics are omitted.\textit{~$^\dagger$ show results from ~\cite{hamamci2025btb3d}}.}
\label{table:radchestct_ood}
\setlength{\tabcolsep}{4.2pt}
\renewcommand{\arraystretch}{1.08}
\begin{tabular}{lccc}
\toprule
Model & Prec.$\uparrow$ & Recall$\uparrow$ & F1$\uparrow$ \\
\midrule
CT2Rep$^{\dagger}$~\cite{CT2Rep}   & 0.299 & 0.139 & 0.133 \\
Merlin$^{\dagger}$~\cite{blankemeier2024merlin}  & 0.271 & 0.149 & 0.182 \\
CT-CHAT$^{\dagger}$~\cite{ctchat} & 0.382 & 0.171 & 0.182 \\
BTB3D-16$^{\dagger}$~\cite{hamamci2025btb3d}& 0.272 & 0.329 & 0.266 \\
\midrule
Base VLM~(Ours)   & 0.443 & 0.418 & 0.395 \\
DCP-PD - DCP1~(Ours)
 & 0.409 & \textbf{0.704} & 0.477 \\
DCP-PD - DCP2~(Ours) & \textbf{0.448} & 0.694 & \textbf{0.503} \\
\bottomrule
\end{tabular}
\end{table}

\noindent\textbf{(RQ3) DCP-PD is both effective and necessary to prevent shortcuts.} 
As shown in Table~\ref{tab:rq3_dropout}, without dropout the GT DCP model collapses when cues are removed at inference, with \textit{No DCP} F1 dropping to $0.094$. In contrast, GT DCP-PD substantially improves robustness to missing cues, increasing \textit{No DCP} F1 from $0.094$ to $0.366$ ($+0.272$).
Further, Table~\ref{tab:attn_reliance} shows consistent evidence from attention. We compute the text/image attention reliance score $S_{\mathrm{text}}$/$S_{\mathrm{image}}$ as defined in Section~\ref{sec:setup}. Under the \textit{DCP-2} test-time setting, the text attention reliance score $S_{\mathrm{text}}$ decreases from $0.441$ (GT DCP) to $0.316$ (GT DCP-PD). This indicates reduced dependence on DCP tokens and a stronger grounding of generation in image tokens.

\begin{figure}[!]
  \centering
\includegraphics[width=\columnwidth,height=4.0cm,keepaspectratio]{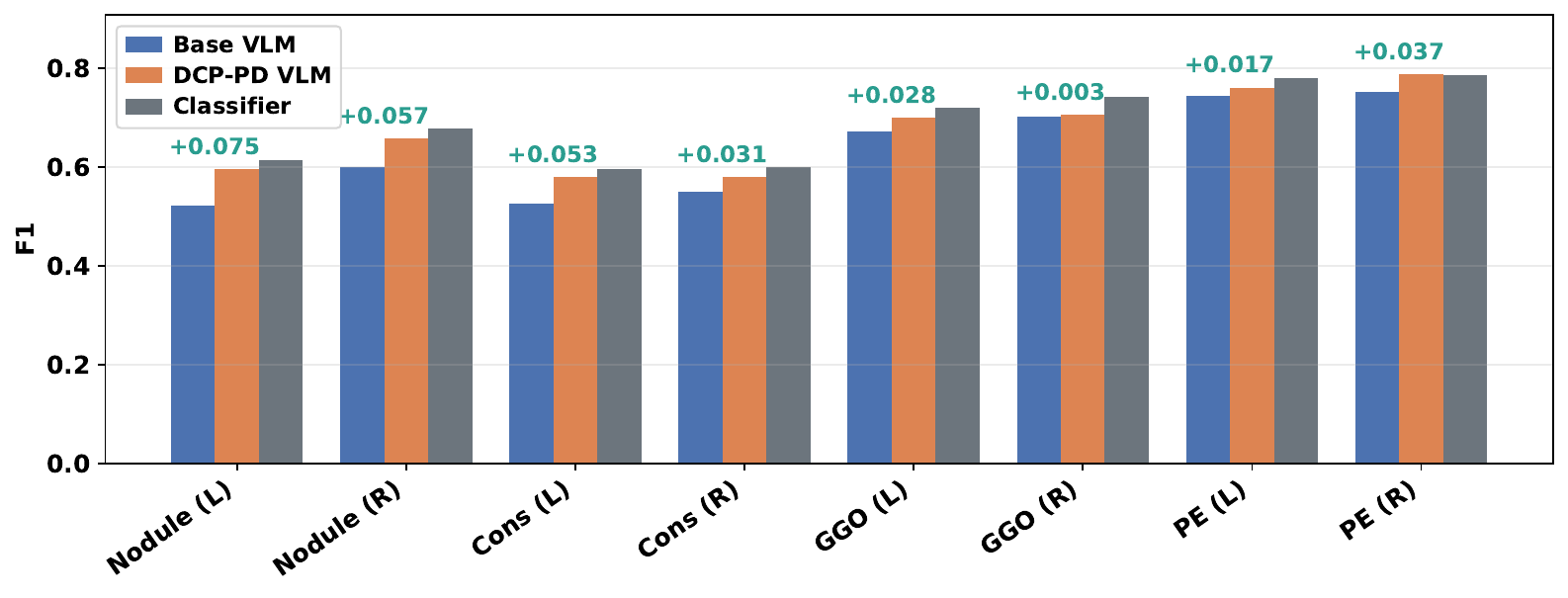}
  \caption{\textbf{Fine-grained evaluation on Laterality.}
We report F1 for the Base VLM, DCP-PD VLM, and trained classifier.
The x-axis includes four findings: lung nodule (Nodule), consolidation (Cons),
ground-glass opacity (GGO), and pleural effusion (PE), each evaluated on
left (L) and right (R) sides.
Numbers above bars indicate $\Delta$F1 (DCP-PD$-$Base).}
  \label{fig:rq5_s1}
\end{figure}

\begin{figure}[!]
  \centering
  \includegraphics[width=\columnwidth]{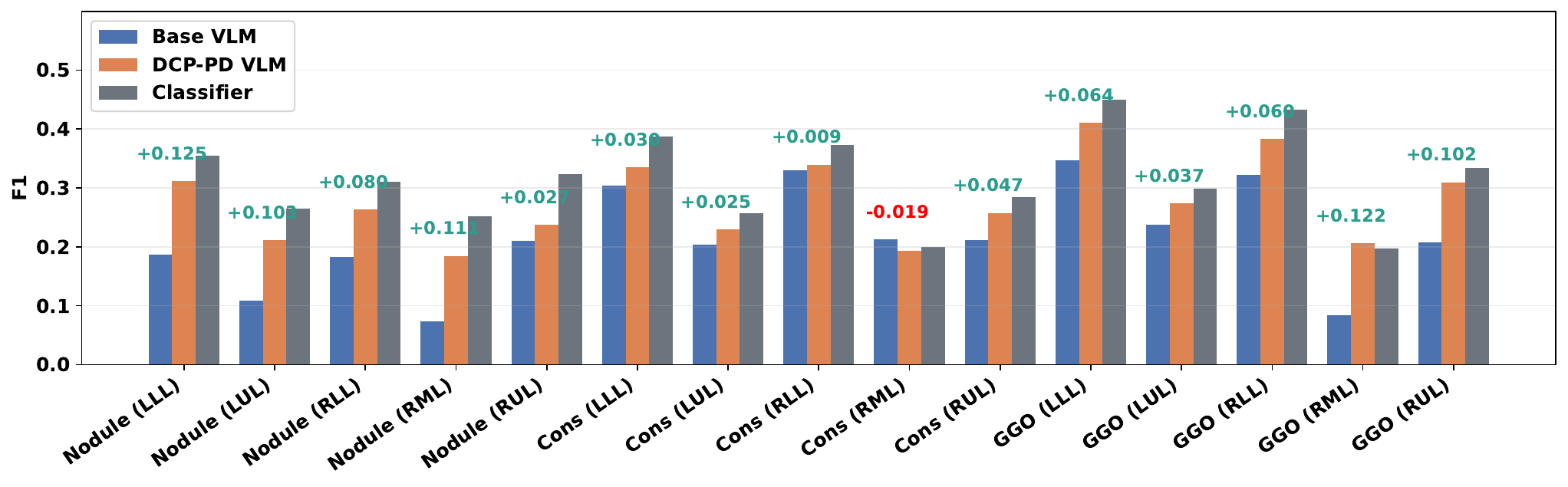}
  \caption{\textbf{Fine-grained evaluation on Lobe.}
  We report F1 for the Base VLM, DCP-PD VLM, and trained classifier on the lobe-level question set.
  Lobe labels correspond to standard lung lobes: left upper lobe (LUL), left lower lobe (LLL), right upper lobe (RUL), right middle lobe (RML), and right lower lobe (RLL).
  Numbers above bars indicate $\Delta$F1 (DCP-PD$-$Base).}
  \label{fig:rq5_s2}
\end{figure}

\noindent\textbf{(RQ4) DCP-PD VLM achieves state-of-the-art performance on in-distribution and out-of-distribution benchmarks.}
As shown in Table~\ref{table:ct_report_results}, on the in-distribution \textbf{CT-RATE} benchmark, our DCP-based model achieves superior clinical performance compared to prior report generation methods, attaining the best F1 and CRG while remaining competitive on lexical metrics. In addition, Fig.~\ref{fig:unstructured_radar} shows that beyond achieving the best average F1, our model attains the highest F1 across all 18 pathology findings.
Furthermore, Table~\ref{table:radchestct_ood} evaluates generalization on the out-of-distribution \textbf{Rad-ChestCT} dataset. Our model maintains state-of-the-art performance on the available metrics compared to prior methods.

\noindent\textbf{(RQ5)} \textbf{DCP-PD readily supports richer cue types and consistently improves performance.}
We extend DCP from finding-level cues to attribute-level cues (laterality and lobe locations) by training additional discriminative models and templating their predictions into cue prompts.
As shown in Fig.~\ref{fig:rq5_s1} and Fig.~\ref{fig:rq5_s2}, DCP-PD consistently improves F1 over the Base VLM across both question sets, narrowing the gap to the standalone classifier and in most cases approaching its performance. One exception is \textit{Cons (RML)} in S2, where DCP-PD underperforms; this aligns with the weaker classifier performance on that entry compared to the Base VLM, limiting the benefit of cue transfer.
Notably, DCP-PD matches or even exceeds the classifier on a few entries (e.g., S1 \textit{PE (R)} and S2 \textit{GGO (RML)}), which points to two potential mechanisms: (i) the model may defer to visual tokens when the cues are incorrect; and (ii) coarse cues may help correct fine-grained localization errors (e.g., \textit{GGO (R)} guiding \textit{GGO (RML)}). Further study is encouraged to disentangle these effects.

\subsection{Ablation Studies}
\label{sec:ablation}

\noindent\textbf{Structured vs.\ unstructured reports.}
We ablate training on structured versus unstructured reports. As shown in Table~\ref{table:struct_id_ood}, structured training consistently improves clinical correctness metrics on both CT-RATE and Rad-ChestCT, with a small drop in lexical overlap on CT-RATE. This is consistent with~\cite{chengrethinking}, who show that narrative-flow training encourages reliance on linguistic priors over visual evidence. Our anatomy-aware structuring similarly reduces cross-sentence dependencies, yielding more focused supervision. The slight drop in lexical scores likely reflects reduced overfitting to stylistic patterns in the training reports.
\begin{table}[ht]
\centering
\caption{\textbf{Effect of report structuring (ID vs.\ OOD).}
We compare the Base VLM trained on unstructured vs.\ structured reports on CT-RATE (ID) and Rad-ChestCT (OOD).
Text-based metrics are unavailable for Rad-ChestCT and are marked as ``--''.}
\label{table:struct_id_ood}
\setlength{\tabcolsep}{2.6pt}
\renewcommand{\arraystretch}{1.05}
\small
\resizebox{\columnwidth}{!}{%
\begin{tabular}{llcccccc}
\toprule
Dataset & Training reports & Prec.$\uparrow$ & Rec.$\uparrow$ & F1$\uparrow$ & CRG$\uparrow$ & BLEU$\uparrow$ & MET$\uparrow$ \\
\midrule
\multirow{2}{*}{CT-RATE (ID)}
& Unstructured & \textbf{0.521} & 0.374 & 0.427 & 0.440 & \textbf{0.323} & \textbf{0.240} \\
& Structured   & 0.520 & \textbf{0.423} & \textbf{0.440} & \textbf{0.457} & 0.293 & 0.214 \\
\midrule
\multirow{2}{*}{Rad-ChestCT (OOD)}
& Unstructured & 0.428 & 0.393 & 0.381 & -- & -- & -- \\
& Structured   & \textbf{0.443} & \textbf{0.418} & \textbf{0.395} & -- & -- & -- \\
\bottomrule
\end{tabular}%
}
\end{table}

\noindent\textbf{Effect of Prompt Dropout Ratio.} We ablate the prompt-dropout ratio $p$ used in DCP-PD training. As shown in Table~\ref{table:ct_report_results_cls}, prompt dropout reliably breaks shortcut reliance, even with a small ratio ($p=0.1$). We also observe a trade-off: increasing $p$ further improves robustness to missing cues, while very large dropout can slightly reduce performance under cue-guided inference.

\begin{table}[ht]
\begin{center}
\caption{Comparison of report generation performance under different classifier-augmented settings.
\textbf{DCP} denotes models trained with a classifier-aware prompt prefix that conditions generation on predicted abnormality labels.
\textbf{GT-guided} represents an oracle setting using ground-truth labels.
\textbf{DCP-2} uses an external classifier's predicted labels, while \textbf{DCP} uses predictions from our trained discriminative model.
\textbf{No DCP} removes the classifier prompt at inference time.
Best results per column are shown in bold.}
\label{table:ct_report_results_cls}
\setlength{\tabcolsep}{3.2pt}
\renewcommand{\arraystretch}{1.05}
\resizebox{\columnwidth}{!}{%
\begin{tabular}{llcccccc}
\toprule
Training setting & Test-time setting
& Prec.$\uparrow$  & Rec.$\uparrow$  & F1$\uparrow$  & CRG$\uparrow$  & BLEU$\uparrow$  & METEOR$\uparrow$ \\
\midrule
\multicolumn{8}{l}{\textbf{Training with GT-DCP}}\\
\midrule
w/ GT DCP & GT-DCP & 0.947 & 0.943 & 0.944 & 0.890 & 0.332 & 0.238 \\
& DCP-2   & 0.534 & 0.716 & 0.606 & 0.561 & 0.294 & 0.231 \\
& DCP-1        & 0.457 & 0.718 & 0.538 &  0.512    &   0.296   &   0.233   \\
\rowcolor{red!10} & No DCP     & 0.393 & 0.071 & 0.094 & 0.349 & 0.268 & 0.200 \\
\midrule
\multicolumn{8}{l}{\textbf{Training with DCP-PD}}\\
\midrule
w/ GT DCP-PD (rate=0.1) & GT-DCP & 0.936 & 0.938 & 0.936 & 0.883 & 0.322 & 0.233 \\
& DCP-2   & 0.531 & 0.712 & 0.602 & 0.560 & 0.299 & 0.228 \\
& DCP-1        & 0.461 & 0.724 & 0.542 &  0.517    &   0.296   &   0.227   \\
& No DCP     & 0.450 & 0.287 & 0.310 & 0.407 & 0.278 & 0.206 \\
\midrule
w/ GT DCP-PD (rate=0.3) & GT-DCP & 0.920 & 0.914 & 0.916 & 0.844 & 0.315 & 0.229 \\
& DCP-2   & 0.540 & 0.700 & 0.603 & 0.555 & 0.306 & 0.226 \\
& DCP-1        & 0.466 & 0.681 & 0.530 &  0.504    &   0.305   &   0.222   \\
& No DCP     & 0.433 & 0.373 & 0.366 & 0.436 & 0.288 & 0.211 \\
\midrule
w/ GT DCP-PD (rate=0.5) & GT-DCP & 0.895 & 0.876 & 0.881 & 0.792 & 0.307 & 0.225 \\
& DCP-2   & 0.534 & 0.668 & 0.585 & 0.540 & 0.306 & 0.222 \\
& DCP-1        & 0.466 & 0.681 & 0.530 &  0.504    &   0.305   &   0.222   \\
& No DCP     & 0.471 & 0.375 & 0.394 & 0.437 & 0.290 & 0.213 \\
\midrule
w/ GT DCP-PD (rate=0.7) & GT-DCP & 0.838 & 0.834 & 0.830 & 0.737 & 0.306 & 0.224 \\
& DCP-2   & 0.543 & 0.649 & 0.581 & 0.535 & 0.306 & 0.222 \\
& DCP-1       & 0.473 & 0.658 & 0.528 &  0.500    &   0.302   &   0.220   \\
& No DCP     & 0.491 & 0.402 & 0.410 & 0.448 & 0.289 & 0.213 \\
\bottomrule
\end{tabular}%
}
\end{center}
\end{table}

\noindent\textbf{Ablation on removing image tokens.}
To understand the role of visual tokens, we perform an ablation where the generator is trained using only discriminative cue prompts without image tokens.
Specifically, the cues consist of the presence/absence of 18 pathology findings (S1-level cues).
We train a cue-conditioned LLM (\textit{DCP-LLM}) using the same training protocol as the full \textit{DCP-PD VLM}, except that visual tokens are removed.
At inference time, both models are guided by predicted cues from DCP-2.

As shown in Table~\ref{tab:no_image_ablation}, \textit{DCP-LLM} performs competitively on standard report-generation benchmarks.
Although it does not use image tokens, it still achieves macro F1 and lexical metrics similar to those of \textit{DCP-PD VLM}.

\begin{figure}[!]
  \centering
  \includegraphics[width=\columnwidth,height=4.0cm,keepaspectratio]{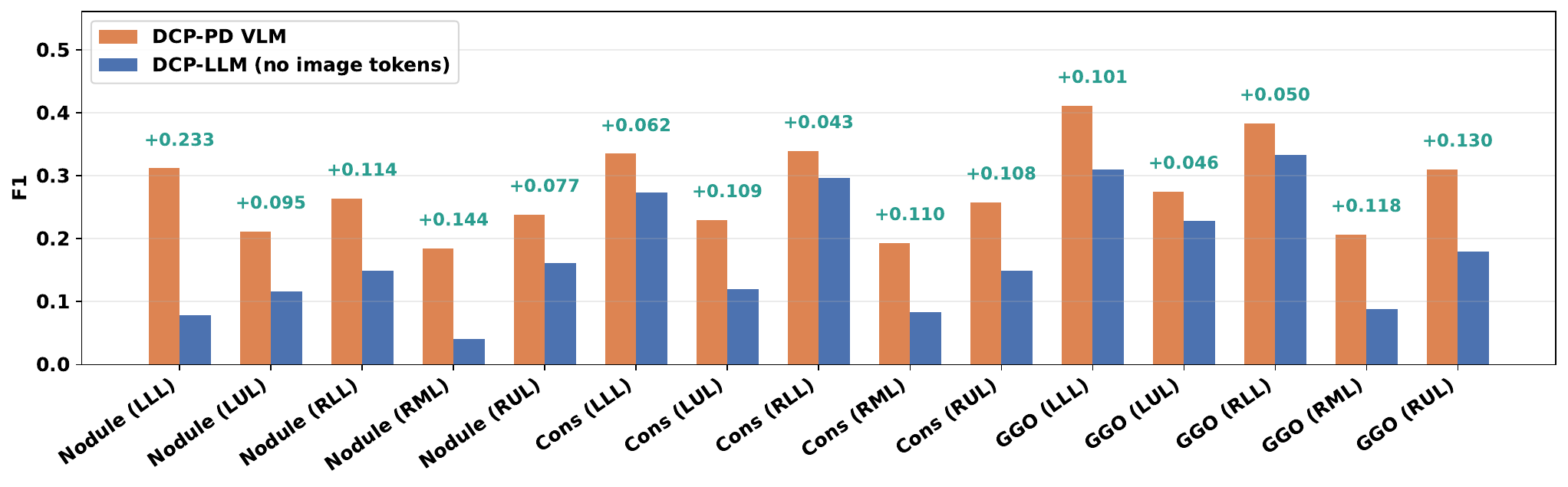}
  \caption{\textbf{Effect of visual tokens on fine-grained lobar grounding.}
We compare the full \textit{DCP-PD VLM} against the cue-only \textit{DCP-LLM} without image tokens.
The x-axis includes three findings: lung nodule (Nodule), consolidation (Cons), and ground-glass opacity (GGO), each evaluated at the lobar level (LLL, LUL, RLL, RML, RUL).
Numbers above bars indicate $\Delta$F1 (\textit{DCP-PD VLM} $-$ \textit{DCP-LLM}), showing that adding visual tokens consistently improves fine-grained spatial grounding.}
  \label{fig:no_image_ablation}
\end{figure}

However, when evaluated on fine-grained spatial grounding, the difference becomes evident.
Figure~\ref{fig:no_image_ablation} shows that \textit{DCP-PD VLM} consistently outperforms the cue-only model on lobar-level localization questions.
Without image tokens, the generator lacks access to spatial visual evidence and therefore struggles to resolve fine-grained locations.

These results highlight two key insights: (i) discriminative cue prompts alone can already drive strong performance on standard report-generation benchmarks, suggesting that such coarse metrics may be partially shortcut by cue-conditioned generation; and (ii) visual tokens remain essential for capturing information beyond the cue space, especially uncued findings and fine-grained spatial details. Together, this explains the advantage of \textit{DCP-PD VLM}: it leverages cues for coarse pathology specification while using image tokens for complementary visual grounding, yielding stronger overall synergy. This also underscores the value of our evaluation protocol, as standard coarse-grained benchmarks can be satisfied by cue-conditioned generation alone, whereas our fine-grained grounding analysis reveals whether the model truly uses image tokens.

\begin{table}[ht]
\centering
\caption{\textbf{Effect of visual tokens in discriminative-cue conditioned report generation on CT-RATE.}
We compare a cue-only generator (\textit{DCP-LLM}, no image tokens) against the full \textit{DCP-PD VLM}. 
We report macro-averaged clinical metrics and report-level generation metrics.}
\label{tab:no_image_ablation}
\setlength{\tabcolsep}{4pt}
\renewcommand{\arraystretch}{1.05}
\begin{tabular}{lccccccc}
\toprule
Model & Prec.$\uparrow$ & Rec.$\uparrow$ & F1$\uparrow$  & CRG$\uparrow$ & BLEU$\uparrow$ & METEOR$\uparrow$ \\
\midrule
DCP-LLM  & 0.531 & \textbf{0.708} & 0.600  & \textbf{0.560} & 0.304 & 0.225 \\
DCP-PD VLM               & \textbf{0.540} & 0.700          & \textbf{0.603}  & 0.555          & \textbf{0.306} & \textbf{0.226} \\
\bottomrule
\end{tabular}
\end{table}

\noindent\textbf{Training-signal ablation: shortcut vs.\ cue neglect.}
We find the training cue signal critically shapes how the generator uses prompt cues.
Training with ground-truth prompts without prompt dropout encourages shortcuts, where the model overly relies on cue tokens (Table~\ref{tab:rq3_dropout}).
Table~\ref{tab:ctrate_dcp_train_test} details the effect of the cue signal used during training.
Although the standalone DCP-1 classifier is strong (Prec.\ 0.439, Rec.\ 0.794, F1 0.544), using its predicted cues for training provides little benefit for report generation: DCP-1-guided training achieves F1 0.438, nearly identical to the Base VLM (0.440).
This suggests that predicted cues provide limited effective supervision, and the generator tends to under-utilize them even when the same cue format is given at inference.
In contrast, training with oracle ground-truth cues yields a much stronger cue-conditioned generator. When evaluated with DCP-1 cues at test time, it achieves an F1 of 0.538, which is close to the standalone classifier performance of 0.544.
Together, these results support our main finding: clean cues can push the VLM close to classifier-level performance, but at the cost of shortcut reliance, whereas noisy predicted cues lead to cue neglect. This trade-off motivates prompt dropout to better balance cue utilization. 

\begin{table}[ht]
\centering
\caption{\textbf{Effect of training/testing cue settings on CT-RATE.}
We report macro-averaged precision, recall, and F1 for the standalone classifier and report generation models under different cue configurations.}
\label{tab:ctrate_dcp_train_test}
\setlength{\tabcolsep}{4.0pt}
\renewcommand{\arraystretch}{1.08}
\begin{tabular}{lllccc}
\toprule
\textbf{Method} & \textbf{Training setting} & \textbf{Testing setting} & \textbf{Prec.$\uparrow$} & \textbf{Rec.$\uparrow$} & \textbf{F1$\uparrow$} \\
\midrule
DCP-1 classifier      & -    & -     & 0.439 & 0.794 & 0.544 \\
Base VLM              & w/o DCP       & w/o DCP  & 0.520 & 0.423 & 0.440 \\
DCP-1-guided training & DCP-1 & DCP-1 & 0.531 & 0.414 & 0.438 \\
GT-guided training    & GT   & DCP-1 & 0.457 & 0.718 & 0.538 \\
\bottomrule
\end{tabular}
\end{table}

\subsection{Visualization}
\begin{figure}[ht]
  \centering
  \includegraphics[width=\columnwidth]{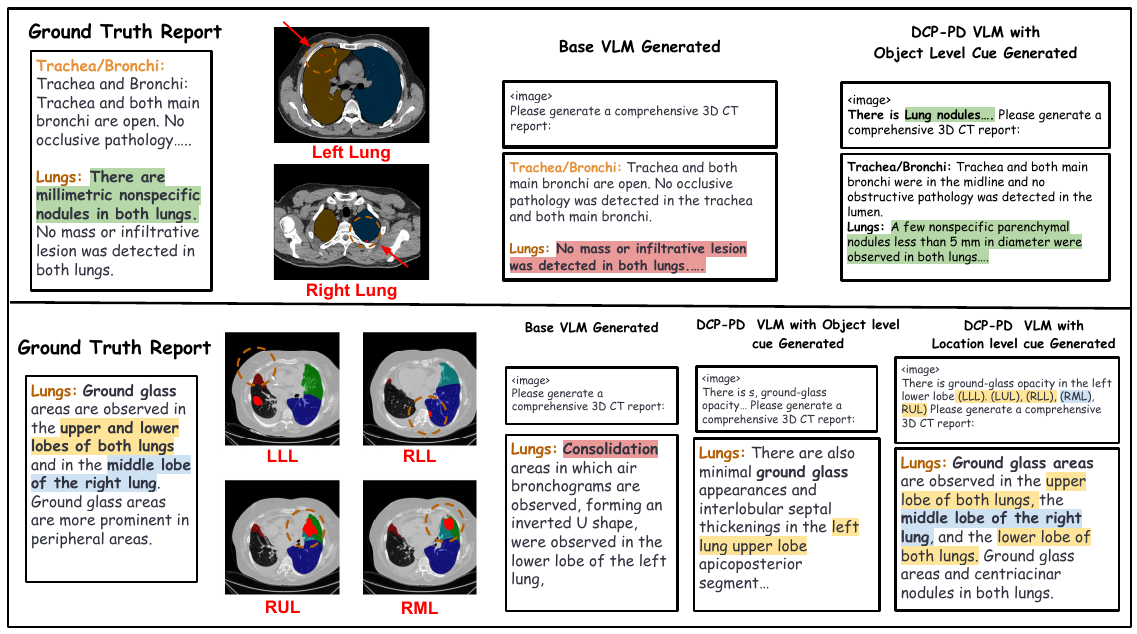}
  \caption{We visualize two examples comparing the Base VLM (no cues) and our DCP-PD model guided at inference by discriminative cue prompts. For each case, we show the ground-truth report (only relevant excerpts due to space), representative CT slices with anatomy overlays from LungMask~\cite{Hofmanninger_2020} and pathology masks from ReXGroundingCT~\cite{baharoon2025rexgroundingct}, and the generated reports. \textbf{Top:} the Base VLM misses the lung nodules, while DCP-PD with object-level cues correctly identifies nodules consistent with the ground-truth report and highlighted regions. \textbf{Bottom:} the Base VLM fails to capture lung opacity and instead reports consolidation; DCP-PD with object-level cues detects the opacity but provides incomplete localization, whereas DCP-PD with fine-grained lobe-level cues produces a more faithful pathology--location description.}
\vspace{-0.8em}

  \label{fig:case_study}
\end{figure}
\begin{figure}[!t]
    \centering
    \includegraphics[width=1\linewidth]{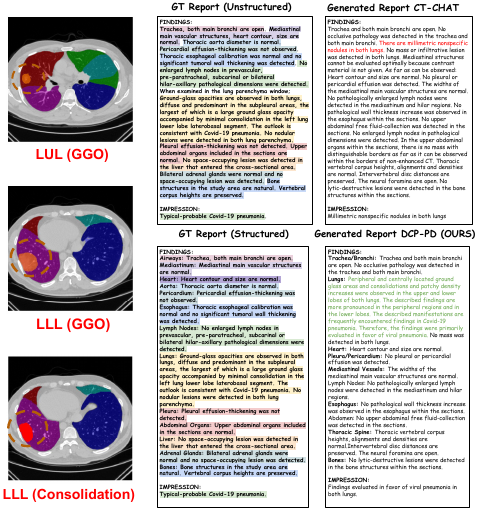}
    \caption{\textbf{Example of full report generation on a CT-RATE validation case.}
The leftmost panel shows representative CT slices with anatomy overlays and pathology masks for GGO and consolidation; abnormal regions are highlighted in red and circled for clarity.
The two report panels on the middle show the original unstructured ground-truth report and our processed structured ground-truth report.
On the right, we compare full-report outputs from a baseline CT-CHAT model trained on unstructured reports and our \textbf{DCP-PD} model conditioned on fine-grained spatial location cues trained on structured reports.
Compared with the baseline, \textbf{DCP-PD} produces a structured report with improved pathology--location grounding.}
    \label{fig:full_report_vis}
\end{figure}


Fig.~\ref{fig:case_study} presents two qualitative examples comparing the Base VLM and our DCP-PD model under different levels of cue guidance. In the first case, the Base VLM misses lung nodules entirely, while DCP-PD with object-level cues correctly identifies them. In the second case, object-level cues alone yield incomplete localization, but fine-grained lobe-level cues produce a more faithful description. Fig.~\ref{fig:full_report_vis} further shows a full report generation comparison between a baseline CT-CHAT model and our DCP-PD model, where DCP-PD produces a structured report with improved pathology--location grounding.

%% file: Sections/main/conclusion.tex
We present \textbf{DCP-PD}, a modular, prompting-based framework for chest CT report generation that integrates discriminative guidance without coupling the \textsc{VLM} to a specific auxiliary model.
By templating discriminative predictions as natural-language cue prompts, our approach is \emph{model-agnostic} and allows the discriminator to be upgraded or replaced at inference time without retraining the underlying \textsc{LLM}.
To mitigate shortcut learning from strong cue supervision, we introduce \emph{prompt dropout}, which reduces reliance on cue tokens and strengthens grounding in visual evidence.
Experiments on CT-RATE and the out-of-distribution Rad-ChestCT benchmark show consistent gains in clinical correctness and state-of-the-art performance on available metrics.
Finally, we demonstrate that DCP-PD naturally scales from finding cues to location cues (laterality and lung lobes) and propose a hierarchical question-set protocol (presence $\rightarrow$ laterality $\rightarrow$ lobe) to more diagnostically evaluate spatial grounding, highlighting that fine-grained localization remains challenging.

%% file: Sections/appendix/appendix_a_v2.tex
\subsection{Multi Scale Attention}
\label{app:msa}
Multi Scale Attention (MSA) builds a multi-scale token hierarchy via simple strided pooling and performs windowed cross-attention to enable bi-directional information flow across scales (fine$\leftrightarrow$coarse). This yields efficient long-context image modeling with $\text{O}(N\log N)$ runtime complexity while preserving global context (details in Atlas~\cite{agrawal2025atlas}).

\subsection{Architecture of DCP-2}
\label{app:dcp2_arch}
DCP-2 uses AnyMC3D~\cite{liu2025revisiting}: a 3D classifier adapted from a frozen 2D foundation model~\cite{oquab2023dinov2}
with LoRA and learnable slice aggregation over slices to output multi-label predictions that are templated into cue prompts.

\subsection{Prompt}
Simple prompt template is used for LLM-based binary question answering.
\label{app:prompt}
\begin{figure}[ht]
\centering
\fbox{
\parbox{0.95\columnwidth}{
\textbf{LLM Parsing Prompt (Single-Question Template)} \\[0.5em]

Based on the following radiology report: \\
\{report\} \\

\{question\} Answer Yes or No only.
}
}
\caption{Simple prompt template used for LLM-based binary question answering.}
\label{fig:prompt_template}
\end{figure}
\newpage
\subsection{Pre-defined Question Sets}
\label{app:question_sets}
\begin{figure}[ht]
\centering
\fbox{
\parbox{0.95\columnwidth}{
\textbf{Finding-level Question Set (Chest CT, 18 findings)} \\[0.5em]

Is there any medical material or device present? \\
Is there arterial wall calcification? \\
Is cardiomegaly or cardiac enlargement suspected based on the imaging findings? \\
Is pericardial effusion present? \\
Is there coronary artery wall calcification? \\
Is there emphysema? \\
Is there any atelectasis? \\
Is there any lung opacity (e.g., ground-glass opacity or other parenchymal opacity)? \\
Is there pulmonary fibrotic sequela? \\
Is there any consolidation in the lung? \\
Is there any lung nodule? \\
Is bronchiectasis present? \\
Is there peribronchial thickening? \\
Is there any mosaic attenuation? \\
Is there any interlobular septal thickening? \\
Is there any pleural effusion? \\
Is there any lymphadenopathy in mediastinum or hila? \\
Is there any hiatal hernia?
}
}
\caption{Finding (QS1) question set used for annotation extraction and report parsing evaluation.}
\label{fig:question_set_18}
\end{figure}

\begin{figure}[ht]
\centering
\fbox{
\parbox{0.95\columnwidth}{
\textbf{Laterality Question Set (Chest CT)} \\[0.5em]

Is there pleural effusion in the right pleural space? \\
Is there pleural effusion in the left pleural space? \\

Is there consolidation in the right lung? \\
Is there consolidation in the left lung? \\

Is there ground-glass opacity in the right lung? \\
Is there ground-glass opacity in the left lung? \\

Is there a lung nodule in the right lung? \\
Is there a lung nodule in the left lung?
}
}
\caption{Laterality (QS2) question set used for annotation extraction and report parsing evaluation.}
\label{fig:question_set_s1}
\end{figure}

\begin{figure}[ht]
\centering
\fbox{
\parbox{0.95\columnwidth}{
\textbf{Lobar Question Set (Chest CT)} \\[0.5em]

Is there consolidation in the right upper lobe (RUL)? \\
Is there consolidation in the right middle lobe (RML)? \\
Is there consolidation in the right lower lobe (RLL)? \\
Is there consolidation in the left upper lobe (LUL)? \\
Is there consolidation in the left lower lobe (LLL)? \\

Is there ground-glass opacity in the right upper lobe (RUL)? \\
Is there ground-glass opacity in the right middle lobe (RML)? \\
Is there ground-glass opacity in the right lower lobe (RLL)? \\
Is there ground-glass opacity in the left upper lobe (LUL)? \\
Is there ground-glass opacity in the left lower lobe (LLL)? \\

Is there a lung nodule in the right upper lobe (RUL)? \\
Is there a lung nodule in the right middle lobe (RML)? \\
Is there a lung nodule in the right lower lobe (RLL)? \\
Is there a lung nodule in the left upper lobe (LUL)? \\
Is there a lung nodule in the left lower lobe (LLL)?
}
}
\caption{Lobar (QS3) question set used for annotation extraction and report parsing evaluation.}
\label{fig:question_set_s2}
\end{figure}
\clearpage

%% file: Sections/appendix/appendix_b.tex
\subsection{Dataset and Preprocessing}
\label{app:data_impl}
\noindent\textbf{Data pre-processing.}
We use the official CT-RATE train/test split, but remove studies corresponding to head-only CT cases, resulting in the exclusion of approximately 800 training samples. For Rad-ChestCT, we use all available data as an external validation set.

\noindent\textbf{CT pre-processing.}
All chest CT volumes are first resampled to a standardized voxel spacing of $1.25\times1.25\times1.25$ mm.
Each CT volume is normalized to 256 axial slices and an in-plane resolution of $256\times256$.
The input is then represented using the raw channel together with multiple tissue-specific CT windows, including lung, mediastinum, abdomen, liver, bone, brain, subdural, stroke, temporal bone, and soft tissue.
Detailed window definitions follow~\cite{pillar0}.
The final input to the Atlas encoder therefore has shape $11\times256\times256\times256$.

\subsection{Training and Implementation Details}
\label{app:train_impl}
\noindent\textbf{Atlas visual backbone.}
The pre-processed CT volume is encoded by the Atlas visual backbone from Pillar-0 with a 3D patchification stem of patch size $(8,8,4)$.
The initial multi-scale hierarchy before encoder-stage updates is given by $\mathbf{h}^{(1)}_{0}$, $\mathbf{h}^{(2)}_{0}$, and $\mathbf{h}^{(3)}_{0}$, with spatial sizes $32\times32\times64$, $4\times4\times64$, and $1\times1\times16$, respectively, and embedding dimension 384 at each scale. For report generation, $\mathbf{h}^{(2)}_{2}$ is used as the visual representation and flattened into a sequence of 1024 tokens with dimension 384.
For discriminative prediction, a multi-scale embedding $\mathbf{h}_{\mathrm{emb}}$ is constructed by pooling over the token dimension at each scale and concatenating the resulting three 384-dimensional vectors, yielding a 1152-dimensional representation.

\noindent\textbf{Visual projector.}
Attention pooling with 256 learnable query tokens of dimension 384 is applied to aggregate the flattened visual tokens.
The resulting pooled features are then projected to the language-token embedding dimension 768 with an MLP.

\noindent\textbf{VLM training.}
The \textsc{VLM} is trained in two stages on 4 H200 GPUs using a per-device batch size of 8.
Stage~1 trains only the vision projector for 1 epoch with the \textsc{LLM} frozen, and takes about 1 hour.
Stage~2 initializes from the Stage~1 checkpoint and performs 5 epochs of supervised fine-tuning with LoRA~\cite{hu2022lora} applied to the \textsc{LLM} (rank 128, scaling factor 256), while jointly optimizing the visual projector; this stage takes about 7 hours.

\noindent\textbf{Discriminative models training.}
Three multi-label linear classifiers are trained on the frozen multi-scale image embedding $\mathbf{h}_{\mathrm{emb}}\in\mathbb{R}^{1152}$, corresponding to the three question sets with 18, 8, and 15 questions, respectively.
Each classifier is implemented as a single fully connected layer, \texttt{nn.Linear}$(1152, C)$, where $C$ is the number of questions in the corresponding set.
Training uses Adam with learning rate $10^{-3}$, batch size 8192, and 1000 epochs.
The loss is class-balanced binary cross-entropy, with per-class positive weights set to $N_{\mathrm{neg}}/N_{\mathrm{pos}}$ computed from the training split.

%% file: Sections/appendix/appendix_c.tex
\subsection{Evaluation Protocol for 18 Pathology Findings}
\label{app:eval_18_pathology}
Following prior work~\cite{ctchat}, we evaluate 18 predefined pathology findings using a RadBERT-based report classifier.
In that work, 1,000 reports were manually annotated for 18 abnormalities by combining the \textit{Findings} and \textit{Impression} sections.
The annotated set was split into 800 training reports and 200 validation reports.
A RadBERT-RoBERTa-4m model~\cite{yan2022radbert}, pretrained on more than 4 million radiology reports, was fine-tuned on this set and then used to classify the remaining reports.

Our question-driven protocol can also support this evaluation.
However, we use the RadBERT-based protocol here for direct comparison with prior work.
DCP-1 is trained with labels from our LLM-based parsing pipeline.
For evaluation, we use pathology labels extracted with RadBERT-RoBERTa-4m to assess both DCP-1 and report generation performance.

On the CT-RATE validation set, our protocol achieves \textbf{0.957} overall agreement with the RadBERT-based evaluation.
This shows strong consistency between the two extraction protocols.
\subsection{Class Distribution of Fine-Grained Annotations}
\label{app:class_distribution}

\begin{table}[ht]
\centering
\small
\caption{Class distribution (Pos/Neg) of the global finding question set (S1) on the CT-RATE validation set.}
\label{tab:posneg_ctrate_valid_s1}
\begin{tabular}{llrr}
\toprule
\textbf{Set} & \textbf{Question} & \textbf{Pos} & \textbf{Neg} \\
\midrule
S1 & Bronchiectasis & 340 & 2699 \\
S1 & Cardiomegaly & 309 & 2730 \\
S1 & Pericardial effusion & 214 & 2825 \\
S1 & Atelectasis & 809 & 2230 \\
S1 & Consolidation & 567 & 2472 \\
S1 & Hiatal hernia & 421 & 2618 \\
S1 & Interlobular septal thickening & 242 & 2797 \\
S1 & Lung nodule & 1372 & 1667 \\
S1 & Lung opacity & 1392 & 1647 \\
S1 & Lymphadenopathy & 218 & 2821 \\
S1 & Medical material / device & 326 & 2713 \\
S1 & Mosaic attenuation pattern & 244 & 2795 \\
S1 & Pleural effusion & 367 & 2672 \\
S1 & Arterial wall calcification & 801 & 2238 \\
S1 & Coronary artery wall calcification & 620 & 2419 \\
S1 & Emphysema & 581 & 2458 \\
S1 & Peribronchial thickening & 361 & 2678 \\
S1 & Pulmonary fibrotic sequela & 401 & 2638 \\
\bottomrule
\end{tabular}

\end{table}

\begin{table}[htbp]
\centering
\small
\caption{Class distribution (Pos/Neg) of the laterality-aware question set (S2) on the CT-RATE validation set.}
\label{tab:posneg_ctrate_valid_s2}
\begin{tabular}{llrr}
\toprule
\textbf{Set} & \textbf{Question} & \textbf{Pos} & \textbf{Neg} \\
\midrule
S2 & Pleural effusion (R) & 303 & 2736 \\
S2 & Pleural effusion (L) & 296 & 2743 \\
S2 & Consolidation (R) & 457 & 2582 \\
S2 & Consolidation (L) & 419 & 2620 \\
S2 & Ground-glass opacity (R) & 883 & 2156 \\
S2 & Ground-glass opacity (L) & 838 & 2201 \\
S2 & Lung nodule (R) & 1144 & 1895 \\
S2 & Lung nodule (L) & 979 & 2060 \\
\bottomrule
\end{tabular}

\end{table}

\begin{table}[htbp]
\centering
\small
\caption{Class distribution (Pos/Neg) of the lobar question set (S3) on the CT-RATE validation set.}
\label{tab:posneg_ctrate_valid_s3}
\begin{tabular}{llrr}
\toprule
\textbf{Set} & \textbf{Question} & \textbf{Pos} & \textbf{Neg} \\
\midrule
S3 & Lung nodule (RUL) & 314 & 2725 \\
S3 & Lung nodule (RML) & 180 & 2859 \\
S3 & Lung nodule (RLL) & 282 & 2757 \\
S3 & Lung nodule (LUL) & 229 & 2810 \\
S3 & Lung nodule (LLL) & 315 & 2724 \\
S3 & Consolidation (RUL) & 143 & 2896 \\
S3 & Consolidation (RML) & 90 & 2949 \\
S3 & Consolidation (RLL) & 224 & 2815 \\
S3 & Consolidation (LUL) & 129 & 2910 \\
S3 & Consolidation (LLL) & 220 & 2819 \\
S3 & Ground-glass opacity (RUL) & 262 & 2777 \\
S3 & Ground-glass opacity (RML) & 128 & 2911 \\
S3 & Ground-glass opacity (RLL) & 393 & 2646 \\
S3 & Ground-glass opacity (LUL) & 245 & 2794 \\
S3 & Ground-glass opacity (LLL) & 413 & 2626 \\
\bottomrule
\end{tabular}

\end{table}

\clearpage

%% file: Sections/appendix/appendix_d_v2.tex
\subsection{Additional Results of DCP1 and DCP2}
\label{app:additional_dcp_results}

\noindent\textbf{CT-RATE.}

\begin{table}[htbp]
\centering
\small
\setlength{\tabcolsep}{4pt}
\caption{Per-question performance of \textbf{DCP-1} on the global finding question set (S1) on the \textbf{CT-RATE validation set}. We report Precision (Prec), Recall (Rec), F1, and AUROC.}
\label{tab:perf_ctrate_valid_s1}
\begin{tabular}{lp{5.8cm}cccc}
\toprule
\textbf{Set} & \textbf{Question} & \textbf{Prec.}$\uparrow$  & \textbf{Rec.}$\uparrow$  & \textbf{F1}$\uparrow$  & \textbf{AUROC}$\uparrow$  \\
\midrule
S1 & Bronchiectasis & 0.266 & 0.688 & 0.383 & 0.789 \\
S1 & Cardiomegaly  & 0.412 & 0.890 & 0.564 & 0.937 \\
S1 & Pericardial effusion & 0.250 & 0.790 & 0.380 & 0.858 \\
S1 & Atelectasis & 0.538 & 0.693 & 0.606 & 0.805 \\
S1 & Consolidation & 0.539 & 0.882 & 0.669 & 0.926 \\
S1 & Hiatal hernia & 0.267 & 0.698 & 0.386 & 0.759 \\
S1 & Interlobular septal thickening & 0.249 & 0.860 & 0.386 & 0.892 \\
S1 & Lung nodule & 0.719 & 0.674 & 0.696 & 0.794 \\
S1 & Lung opacity & 0.829 & 0.757 & 0.791 & 0.885 \\
S1 & Lymphadenopathy & 0.237 & 0.821 & 0.368 & 0.883 \\
S1 & Medical material / device & 0.462 & 0.831 & 0.594 & 0.934 \\
S1 & Mosaic attenuation pattern & 0.270 & 0.762 & 0.398 & 0.873 \\
S1 & Pleural effusion & 0.774 & 0.935 & 0.847 & 0.982 \\
S1 & Arterial wall calcification & 0.655 & 0.858 & 0.743 & 0.914 \\
S1 & Coronary artery wall calcification & 0.597 & 0.900 & 0.718 & 0.930 \\
S1 & Emphysema & 0.402 & 0.673 & 0.504 & 0.811 \\
S1 & Peribronchial thickening & 0.279 & 0.715 & 0.402 & 0.814 \\
S1 & Pulmonary fibrotic sequela & 0.249 & 0.599 & 0.351 & 0.706 \\
\midrule
 & \textbf{Average} & \textbf{0.444} & \textbf{0.779} & \textbf{0.544} & \textbf{0.861} \\
\bottomrule
\end{tabular}
\end{table}

\begin{table}[htbp]
\centering
\small
\setlength{\tabcolsep}{4pt}
\caption{Per-question performance of\textbf{ DCP-2 }on the\textbf{ CT-RATE validation set }for the global finding question set (S1). We report Precision (Prec), Recall (Rec), F1, and AUROC.}
\label{tab:perf_ctrate_valid_s1_dcp2}
\begin{tabular}{lp{5.8cm}cccc}
\toprule
\textbf{Set} & \textbf{Question} & \textbf{Prec.}$\uparrow$  & \textbf{Rec.}$\uparrow$  & \textbf{F1}$\uparrow$  & \textbf{AUROC}$\uparrow$  \\
\midrule
S1 & Medical material / device & 0.545 & 0.741 & 0.628 & 0.932 \\
S1 & Arterial wall calcification & 0.685 & 0.943 & 0.793 & 0.946 \\
S1 & Cardiomegaly & 0.608 & 0.686 & 0.645 & 0.947 \\
S1 & Pericardial effusion & 0.571 & 0.659 & 0.612 & 0.920 \\
S1 & Coronary artery wall calcification & 0.709 & 0.916 & 0.799 & 0.952 \\
S1 & Hiatal hernia & 0.525 & 0.571 & 0.547 & 0.847 \\
S1 & Lymphadenopathy & 0.517 & 0.676 & 0.586 & 0.788 \\
S1 & Emphysema & 0.490 & 0.662 & 0.563 & 0.836 \\
S1 & Atelectasis & 0.440 & 0.805 & 0.569 & 0.819 \\
S1 & Lung nodule & 0.488 & 0.977 & 0.651 & 0.774 \\
S1 & Lung opacity & 0.634 & 0.890 & 0.741 & 0.883 \\
S1 & Pulmonary fibrotic sequela & 0.423 & 0.771 & 0.546 & 0.752 \\
S1 & Pleural effusion & 0.840 & 0.851 & 0.845 & 0.967 \\
S1 & Mosaic attenuation pattern& 0.521 & 0.581 & 0.550 & 0.928 \\
S1 & Peribronchial thickening & 0.373 & 0.600 & 0.460 & 0.841 \\
S1 & Consolidation & 0.569 & 0.850 & 0.682 & 0.924 \\
S1 & Bronchiectasis & 0.414 & 0.633 & 0.501 & 0.869 \\
S1 & Interlobular septal thickening & 0.385 & 0.687 & 0.494 & 0.911 \\
 & \textbf{Average} & \textbf{0.541} & \textbf{0.750} & \textbf{0.623} & \textbf{0.880} \\

\bottomrule
\end{tabular}
\end{table}

\begin{table}[htbp]
\centering
\small
\setlength{\tabcolsep}{4pt}
\caption{Per-question performance of \textbf{DCP-1 }on the laterality-aware question set (S2) on the \textbf{CT-RATE validation set}. We report Precision (Prec), Recall (Rec), F1, and AUROC.}
\label{tab:perf_ctrate_valid_s2}
\begin{tabular}{lp{5.8cm}cccc}
\toprule
\textbf{Set} & \textbf{Question} & \textbf{Prec.}$\uparrow$  & \textbf{Rec.}$\uparrow$  & \textbf{F1}$\uparrow$  & \textbf{AUROC}$\uparrow$  \\
\midrule
S2 & Pleural effusion (R) & 0.680 & 0.931 & 0.786 & 0.982 \\
S2 & Pleural effusion (L) & 0.667 & 0.936 & 0.779 & 0.978 \\
S2 & Consolidation (R) & 0.461 & 0.860 & 0.600 & 0.922 \\
S2 & Consolidation (L) & 0.445 & 0.905 & 0.596 & 0.923 \\
S2 & Ground-glass opacity (R) & 0.687 & 0.804 & 0.741 & 0.898 \\
S2 & Ground-glass opacity (L) & 0.655 & 0.796 & 0.719 & 0.893 \\
S2 & Lung nodule (R) & 0.641 & 0.718 & 0.677 & 0.800 \\
S2 & Lung nodule (L) & 0.544 & 0.702 & 0.613 & 0.779 \\
\midrule
 & \textbf{Average} & \textbf{0.598} & \textbf{0.832} & \textbf{0.689} & \textbf{0.897} \\
\bottomrule
\end{tabular}
\end{table}

\begin{table}[htbp]
\centering
\small
\setlength{\tabcolsep}{4pt}
\caption{Per-question performance of \textbf{DCP-1 }on the lobar question set (S3) on the \textbf{CT-RATE validation set}. We report Precision (Prec), Recall (Rec), F1, and AUROC.}
\label{tab:perf_ctrate_valid_s3}
\begin{tabular}{lp{5.8cm}cccc}
\toprule
\textbf{Set} & \textbf{Question} & \textbf{Prec.}$\uparrow$ & \textbf{Rec.}$\uparrow$ & \textbf{F1}$\uparrow$ & \textbf{AUROC}$\uparrow$ \\
\midrule
S3 & Lung nodule (RUL) & 0.214 & 0.666 & 0.324 & 0.768 \\
S3 & Lung nodule (RML) & 0.154 & 0.694 & 0.252 & 0.795 \\
S3 & Lung nodule (RLL) & 0.201 & 0.684 & 0.311 & 0.762 \\
S3 & Lung nodule (LUL) & 0.166 & 0.664 & 0.265 & 0.747 \\
S3 & Lung nodule (LLL) & 0.239 & 0.686 & 0.355 & 0.783 \\
S3 & Consolidation (RUL) & 0.172 & 0.818 & 0.285 & 0.903 \\
S3 & Consolidation (RML) & 0.113 & 0.856 & 0.200 & 0.896 \\
S3 & Consolidation (RLL) & 0.240 & 0.830 & 0.373 & 0.895 \\
S3 & Consolidation (LUL) & 0.150 & 0.876 & 0.257 & 0.897 \\
S3 & Consolidation (LLL) & 0.249 & 0.868 & 0.387 & 0.905 \\
S3 & Ground-glass opacity (RUL) & 0.212 & 0.790 & 0.334 & 0.811 \\
S3 & Ground-glass opacity (RML) & 0.113 & 0.789 & 0.197 & 0.825 \\
S3 & Ground-glass opacity (RLL) & 0.302 & 0.763 & 0.433 & 0.818 \\
S3 & Ground-glass opacity (LUL) & 0.188 & 0.727 & 0.298 & 0.798 \\
S3 & Ground-glass opacity (LLL) & 0.317 & 0.772 & 0.450 & 0.820 \\
\midrule
 & \textbf{Average} & \textbf{0.202} & \textbf{0.766} & \textbf{0.315} & \textbf{0.828} \\
\bottomrule
\end{tabular}
\end{table}

\clearpage
\noindent\textbf{Rad-Chest CT.} Due to annotation mismatch between CT-RATE and Rad-ChestCT, we follow prior work~\cite{hamamci2025btb3d} and report 16-label predictions derived from the original 18-label outputs. S1$^{*}$ denotes the 16-label evaluation protocol on Rad-ChestCT derived from the original 18-label prediction space.
\begin{table}[htbp]
\centering
\small
\setlength{\tabcolsep}{4pt}
\caption{Per-question performance of \textbf{DCP-1} on \textbf{Rad-ChestCT }for the global finding question set (S1$^{*}$). We report Precision (Prec), Recall (Rec), F1, and AUROC.}
\label{tab:perf_radchest_s1_dcp1}
\begin{tabular}{lp{5.8cm}cccc}
\toprule
\textbf{Set} & \textbf{Question} & \textbf{Prec.}$\uparrow$  & \textbf{Rec.}$\uparrow$  & \textbf{F1}$\uparrow$  & \textbf{AUROC}$\uparrow$  \\
\midrule
S1$^{*}$ & Medical material & 0.489 & 0.896 & 0.633 & 0.834 \\
S1$^{*}$ & Calcifications & 0.838 & 0.879 & 0.858 & 0.802 \\
S1$^{*}$ & Cardiomegaly & 0.341 & 0.759 & 0.470 & 0.870 \\
S1$^{*}$ & Pericardial effusion & 0.255 & 0.744 & 0.380 & 0.758 \\
S1$^{*}$ & Hiatal hernia & 0.238 & 0.609 & 0.342 & 0.745 \\
S1$^{*}$ & Lymphadenopathy & 0.275 & 0.855 & 0.416 & 0.807 \\
S1$^{*}$ & Emphysema & 0.421 & 0.957 & 0.585 & 0.910 \\
S1$^{*}$ & Atelectasis & 0.398 & 0.832 & 0.538 & 0.732 \\
S1$^{*}$ & Lung nodule & 0.913 & 0.735 & 0.814 & 0.802 \\
S1$^{*}$ & Lung opacity & 0.680 & 0.609 & 0.642 & 0.694 \\
S1$^{*}$ & Pulmonary fibrotic sequela & 0.196 & 0.856 & 0.319 & 0.738 \\
S1$^{*}$ & Pleural effusion & 0.518 & 0.945 & 0.669 & 0.956 \\
S1$^{*}$ & Peribronchial thickening & 0.125 & 0.763 & 0.214 & 0.698 \\
S1$^{*}$ & Consolidation & 0.313 & 0.822 & 0.454 & 0.834 \\
S1$^{*}$ & Bronchiectasis & 0.214 & 0.887 & 0.344 & 0.736 \\
S1$^{*}$ & Interlobular septal thickening & 0.164 & 0.825 & 0.273 & 0.826 \\
\midrule
 & \textbf{Average} & \textbf{0.399} & \textbf{0.811} & \textbf{0.497} & \textbf{0.796} \\
\bottomrule
\end{tabular}
\end{table}

\begin{table}[htbp]
\centering
\small
\setlength{\tabcolsep}{4pt}
\caption{Per-question performance of \textbf{DCP-2} on \textbf{Rad-ChestCT} for the global finding question set (S1$^{*}$). We report Precision (Prec), Recall (Rec), F1, and AUROC. Mosaic attenuation pattern is omitted because no positive samples are present in this evaluation set.}
\label{tab:perf_radchest_s1_dcp2}
\begin{tabular}{lp{5.8cm}cccc}
\toprule
\textbf{Set} & \textbf{Question} & \textbf{Prec.}$\uparrow$  & \textbf{Rec.}$\uparrow$  & \textbf{F1}$\uparrow$  & \textbf{AUROC}$\uparrow$  \\
\midrule
S1$^{*}$ & Medical material & 0.615 & 0.706 & 0.658 & 0.814 \\
S1$^{*}$ & Calcifications & 0.803 & 0.920 & 0.857 & 0.818 \\
S1$^{*}$ & Cardiomegaly & 0.399 & 0.716 & 0.512 & 0.897 \\
S1$^{*}$ & Pericardial effusion & 0.586 & 0.538 & 0.561 & 0.849 \\
S1$^{*}$ & Hiatal hernia & 0.368 & 0.668 & 0.475 & 0.828 \\
S1$^{*}$ & Lymphadenopathy & 0.205 & 0.858 & 0.331 & 0.729 \\
S1$^{*}$ & Emphysema & 0.391 & 0.970 & 0.558 & 0.932 \\
S1$^{*}$ & Atelectasis & 0.358 & 0.927 & 0.517 & 0.732 \\
S1$^{*}$ & Lung nodule & 0.823 & 0.973 & 0.892 & 0.744 \\
S1$^{*}$ & Lung opacity & 0.621 & 0.801 & 0.700 & 0.680 \\
S1$^{*}$ & Pulmonary fibrotic sequela & 0.119 & 0.743 & 0.206 & 0.422 \\
S1$^{*}$ & Pleural effusion & 0.800 & 0.794 & 0.797 & 0.943 \\
S1$^{*}$ & Peribronchial thickening & 0.159 & 0.711 & 0.259 & 0.733 \\
S1$^{*}$ & Consolidation & 0.343 & 0.634 & 0.445 & 0.801 \\
S1$^{*}$ & Bronchiectasis & 0.289 & 0.835 & 0.430 & 0.824 \\
S1$^{*}$ & Interlobular septal thickening & 0.202 & 0.825 & 0.324 & 0.855 \\
\midrule
 & \textbf{Average} & \textbf{0.443} & \textbf{0.789} & \textbf{0.533} & \textbf{0.788} \\
\bottomrule
\end{tabular}
\end{table}